\documentclass[10pt,twocolumn,letterpaper]{article}

\usepackage[pagenumbers]{cvpr} 

\usepackage{graphicx}
\usepackage{amsmath}
\usepackage{amssymb}
\usepackage{booktabs}
\usepackage{times}
\usepackage{epsfig}
\usepackage{graphicx}
\usepackage{amsmath}
\usepackage{amssymb}
\usepackage{color,soul}
\usepackage{graphicx}
\usepackage{caption}
\usepackage{subcaption}
\usepackage{multirow}
\usepackage{makecell}
\usepackage[ruled,vlined]{algorithm2e}
\usepackage{silence}
\usepackage{multirow}
\usepackage{float}
\usepackage{makecell}
\usepackage{bbm}
\usepackage{amsmath, amssymb, mathtools}
\usepackage[accsupp]{axessibility}

\usepackage[pagebackref,breaklinks,colorlinks]{hyperref}

\usepackage[capitalize]{cleveref}
\crefname{section}{Sec.}{Secs.}
\Crefname{section}{Section}{Sections}
\Crefname{table}{Table}{Tables}
\crefname{table}{Tab.}{Tabs.}

\def\cvprPaperID{9541} 
\def\confName{CVPR}
\def\confYear{2022}

\makeatletter
\newcommand{\printfnsymbol}[1]{%
  \textsuperscript{\@fnsymbol{#1}}%
}
\newcommand{\printfnsymbolnew}[2]{%
  \textsuperscript{\@fnsymbol{#1}}%
}
\makeatother

\begin{document}

%

\title{Investigating the Impact of Multi-LiDAR Placement  on \\ Object Detection   for Autonomous Driving}

\author{Hanjiang Hu$^{1}$\thanks{equal contribution}\quad  Zuxin Liu$^{1}$\printfnsymbol{1}\,\, Sharad Chitlangia$^{2}$\thanks{work done while interning at CMU, prior to current affiliations}\quad  Akhil Agnihotri$^{3}$\printfnsymbolnew{2}\,\, Ding Zhao$^{1}$\\
$^1$Carnegie Mellon University\quad   $^2$Amazon\quad  $^3$University of Southern California
\\ {\tt\small \{hanjianghu,zuxinl,dingzhao\}@cmu.edu, chitshar@amazon.com, akhil.agnihotri@usc.edu}
}
\maketitle

\begin{abstract}
   The past few years have witnessed an increasing interest in improving the perception performance of LiDARs on autonomous vehicles. While most of the existing works focus on developing new deep learning algorithms or model architectures, we study the problem from the physical design perspective, \textit{i.e.}, how different placements of multiple LiDARs influence the learning-based perception. To this end, we introduce an easy-to-compute information-theoretic surrogate metric to  quantitatively and fast evaluate LiDAR placement for 3D detection of different types of objects. We also present a new data collection, detection model training and evaluation framework in the realistic CARLA simulator to evaluate disparate multi-LiDAR configurations. Using several prevalent placements inspired by the designs of self-driving companies, we show the correlation between our surrogate metric and object detection performance of different representative algorithms on KITTI through extensive experiments,
   validating the effectiveness of our LiDAR placement evaluation  approach.
   Our results show that sensor placement is non-negligible in 3D point cloud-based object detection, which will contribute up to $10\%$ performance discrepancy in terms of average precision in challenging 3D object detection settings.
   We believe that this is one of the first studies to quantitatively investigate the influence of LiDAR placement on perception performance.
\end{abstract}

\section{Introduction}
\label{sec:intro}
LiDAR sensors are becoming the critical 3D sensors for autonomous vehicles (AVs) since they could provide accurate 3D geometry information and precise distance measures under various driving conditions \cite{liu2019lpd, 9635878}. The point cloud data generated from LiDARs has been used to perform a series of perception tasks, such as object detection and tracking \cite{shi2021pv, shi2020points, shi2019pointrcnn}, SLAM and localization \cite{zhang2014loam,lu2019deepvcp,yu2018ds}.
\begin{figure}[h]
\begin{center}
    \begin{subfigure}[b]{0.22\textwidth}
                \includegraphics[width=\linewidth]{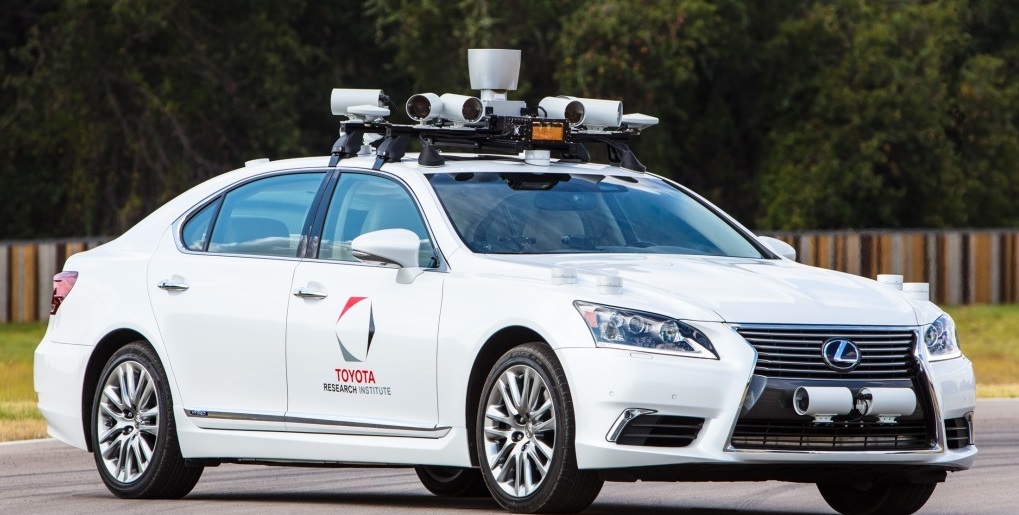}
                \caption{Toyota}
                \label{fig:toyota}
    \end{subfigure}
    \begin{subfigure}[b]{0.22\textwidth}
                \includegraphics[width=\linewidth]{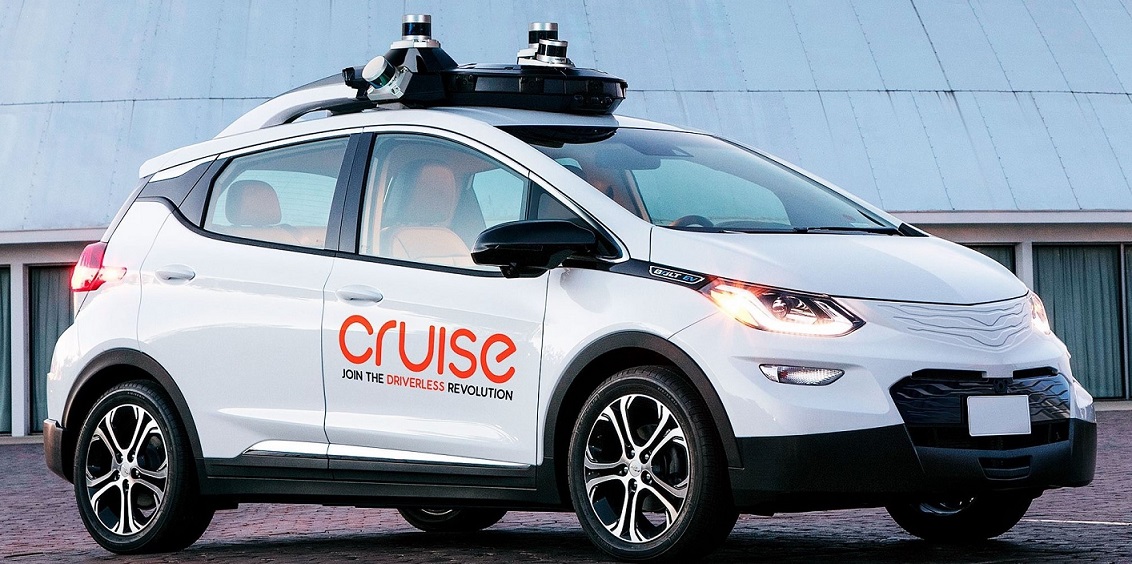}
                \caption{Cruise}
                \label{fig:cruise}
    \end{subfigure}

    \begin{subfigure}[b]{0.22\textwidth}
                \includegraphics[width=\linewidth]{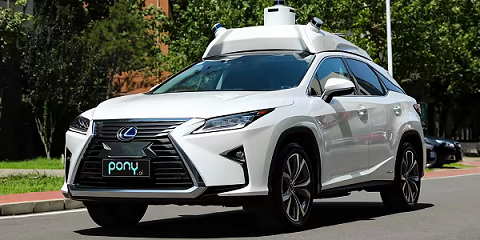}
                \caption{Pony.ai}
                \label{fig:ponyai}
    \end{subfigure}
    \begin{subfigure}[b]{0.22\textwidth}
                \includegraphics[width=\linewidth,height=0.5\linewidth]{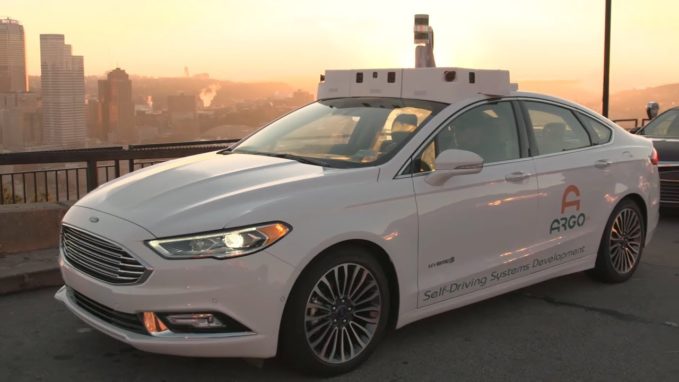}
                \caption{Argo AI}
                \label{fig:argoai}
    \end{subfigure}%

    \begin{subfigure}[b]{0.22\textwidth}
                \includegraphics[width=\linewidth,height=0.61\linewidth]{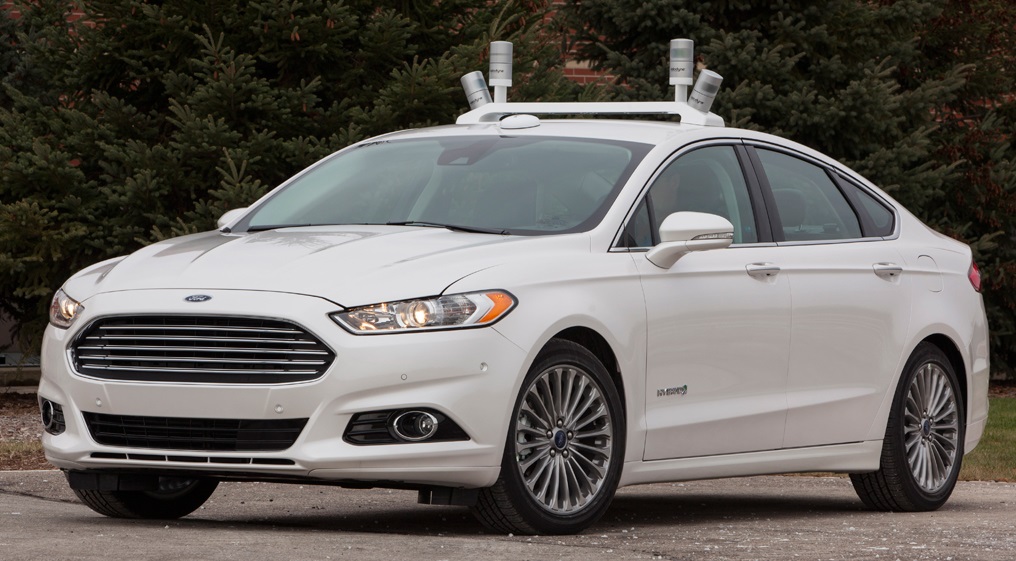}
                \caption{Ford}
                \label{fig:ford}
    \end{subfigure}
    \begin{subfigure}[b]{0.22\textwidth}
                \includegraphics[width=\linewidth,height=0.61\linewidth]{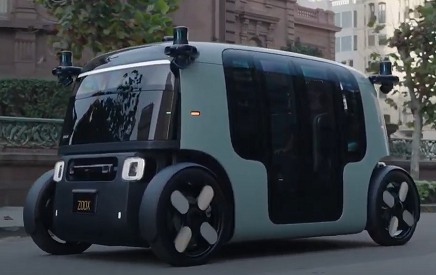}
                \caption{Zoox}
                \label{fig:zoox}
    \end{subfigure}
\end{center}
\vspace{-8mm}
  \caption{Different multi-LiDAR configurations used in different autonomous vehicles \cite{toyota,cruise,ponyai,argoai,ford,zoox}.}
\label{fig:baseline-config}
\vspace{-6mm}
\end{figure}

High-quality point cloud data  and effective perception algorithms are usually both required to achieve accurate 3D object detection in practice \cite{xu2021spg}. A number of studies propose to improve the 3D object detection performance for point cloud data by developing novel perception algorithms, which assume that the data is of high quality~\cite{shi2020points,shi2019pointrcnn, lang2019pointpillars, yan2018second, deng2020voxel}. However, only a little literature considers the LiDAR perception problem from the viewpoint of data acquisition \cite{liu2019should,kini2020sensor,ma2021perception} and LiDAR placement. We believe that this new perspective should be equally crucial for real-world AV applications since improper LiDAR placement may cause poor-quality sensing data, and thus corrupt the perception algorithm and lead to poor performance \cite{cao2019adversarial,liu2019extending, xu2021spg, kini2020sensor,liu2019should,mou2018optimal}. In addition, LiDAR is an expensive sensor, so maximizing its utility to save the deployment cost is also important for mass production.
Therefore, we aim to investigate the interplay between LiDAR sensor placement and perception performance for AVs. We use placement and configuration interchangeably throughout this paper.

\begin{figure*}[t!]
\begin{center}
 \includegraphics[width=0.8\linewidth]{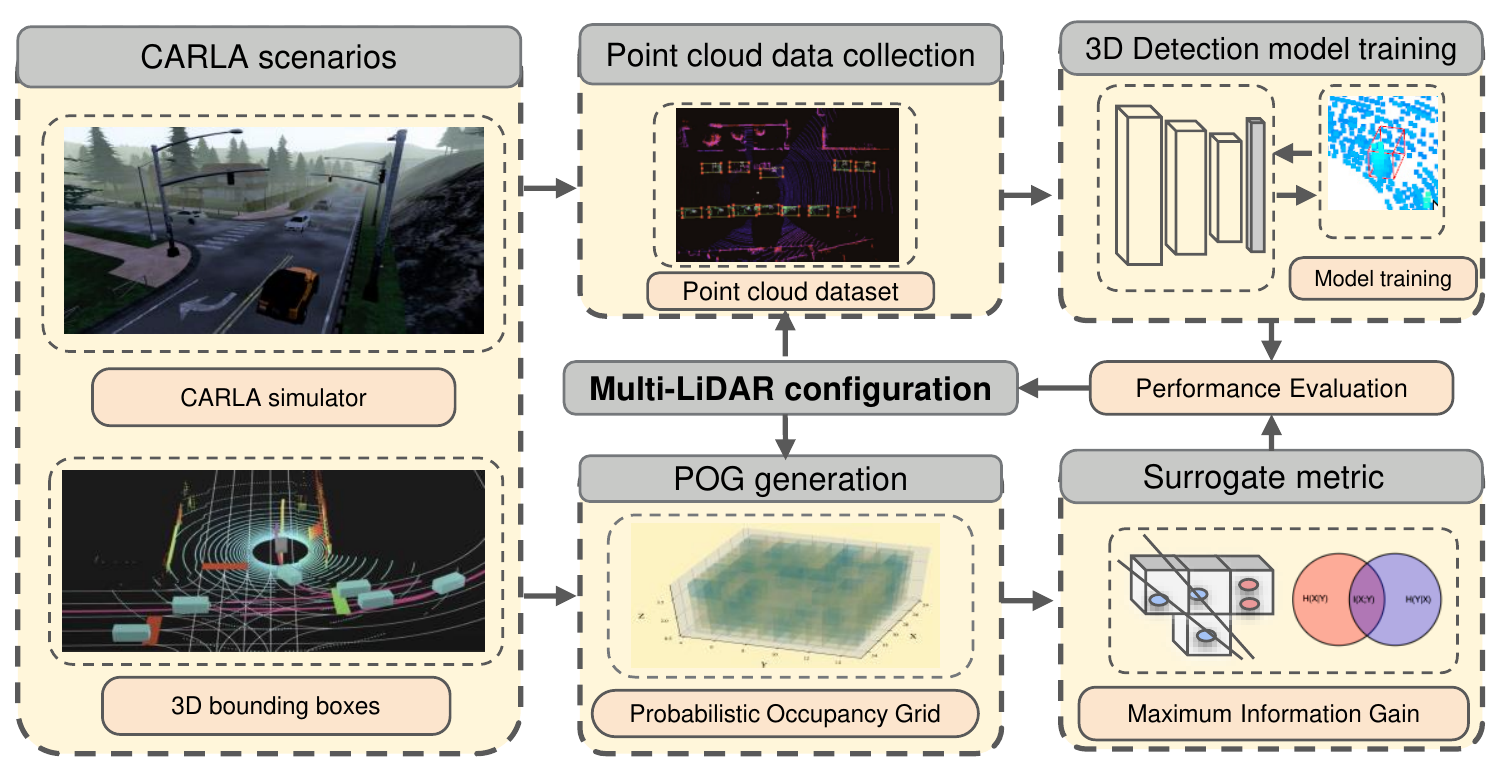}
\end{center}
\vspace*{-5mm}
  \caption{Evaluation framework overview.}
\label{fig:overview}
\vspace*{-5mm}
\end{figure*}

However, it is not easy to evaluate the efficacy of different LiDAR placement layouts based on the perception performance in the real world, which comes with effort-costly and time-consuming full loops of LiDAR deployment, data collection, model training, and performance evaluation.
Moreover, as shown in Figure \ref{fig:baseline-config}, many companies' self-driving vehicles are equipped with more than 2 LiDARs. As the LiDAR number increases, the cost of LiDAR configuration evaluation and optimization will also increase exponentially.
Therefore, it is a crucial but still open problem to accelerate the quantitative evaluation of the LiDAR configurations regarding perception performance with low cost. Thoroughly studying the interplay between LiDAR sensor placement and perception performance is essential to AV perception systems, saving deployment cost and without sacrificing driving safety.
In this paper, we study the perception system from the sensing perspective and focus on investigating the relationship between LiDAR configurations and 3D object detection using our evaluation framework shown in Figure \ref{fig:overview}.
The contributions of this paper are summarized as follows:
\begin{itemize}
    \item We establish a systematic framework to evaluate the object detection performance of different LiDAR placements and investigate prevalent LiDAR placements inspired by self-driving companies, showing that LiDAR placement dramatically influences the performance of object detection up to 10\% in challenging 3D detection settings. As far as we know, we are one of the first works to quantitatively study the interplay between LiDAR placement and perception performance.

    \item We propose a novel surrogate metric with maximum information gain (S-MIG) to accelerate the evaluation of LiDAR placement by modeling object distribution through the proposed scalable Probabilistic Occupancy Grids (POG). We show the correlation between the surrogate metric and detection performance, which is theoretically explained via S-MIG and validates its effectiveness for the LiDAR configuration evaluation.

    \item  We contribute an automated multi-LiDAR data collection and detection model evaluation pipeline in the realistic CARLA simulator and conduct extensive experiments with state-of-the-art LiDAR-based 3D object detection algorithms. The results reveal that LiDAR placement plays an important role in the perception system. The code for the framework is available on \url{https://github.com/HanjiangHu/Multi-LiDAR-Placement-for-3D-Detection}.


\end{itemize}


\section{Related Work}


\label{related-work}
The methods and frameworks proposed in this work revolve around evaluating the LiDAR sensor placement for point cloud-based 3D object detection in autonomous driving. Although the literature is scarce in this combined area, there has been some research on the 3D object detection and the LiDAR placement optimization topics independently, which we discuss in this section.

\textbf{LiDAR-based 3D object detection.} To keep up with the surge  in the LiDAR applications in autonomous vehicles, researchers have tried to develop novel point cloud-based 3D object detection from LiDAR. Over the years, there has also been great progress in grid-based and point-based 3D detection methods for point cloud data. 2D grid-based methods project the point cloud data onto a 2D bird-eye-view (BEV) to generate bounding boxes \cite{chen2017multi, liang2019multi, vora2020pointpainting, vora2020pointpainting, Yin_2021_CVPR, lang2019pointpillars}. Another way to deal with point cloud is to use 3D voxels and 3D CNN, which is also called voxel-based methods \cite{zhou2018voxelnet,  yan2018second, shi2020points, graham20183d, deng2020voxel}. However, these grid-based methods are greatly limited by the kernel size of the 2D or 3D convolution, although the region proposal is pretty accurate with fast detection speed. Point-based methods, on the other hand, directly apply on the raw point cloud based on PointNet series \cite{qi2017pointnet, qi2017pointnet++, qi2018frustum}, which enables them to have flexible perception leading to robust point cloud learning \cite{wang2019dynamic,yang2019std, qi2019deep, yang20203dssd}.  More receptive fields bring about higher computation costs compared to grid-based methods. Recent work  \cite{shi2020pv, shi2021pv} combine voxel-based and point-based methods together to efficiently learn features from raw point cloud data, leading to impressive results on KITTI dataset \cite{geiger2013vision}. To evaluate the influence of point cloud of LiDAR configuration, we use some representative voxel-based and point-based methods \cite{openpcdet2020} in Section \ref{experiments}. The majority of the detection methods mentioned above are well-designed and evaluated on high-quality point cloud datasets, but do not consider the influence of the LiDAR sensing system.

\textbf{LiDAR placement for autonomous vehicles.}
One of the critical factors for autonomous vehicles is the perception and sensing ability. With this respect, LiDARs have been widely used because of their high real-time precision and their ability to extract rich  information from the environment \cite{maddern2014illumination, geiger2013vision, zhang2014loam}.  However, the perception ability of the LiDAR is sensitive to its placement \cite{zhang1995two, durrant1987consistent}, so it is critical to developing a scheme that minimizes the uncertainty among all its possible placements.
To this extent, Dybedal \etal \cite{dybedal2017optimal} proposed to find the optimal placement of 3D sensors using a Mixed Integer Linear Programming approach, which is not scalable to AVs because of the large number of variables involved. Rahimian \etal  \cite{rahimian2016optimal} developed a dynamic occlusion-based optimal placement routine for 3D motion capture systems, but do not consider variable number of sensors during its optimization routine.

There have also been some prominent advances to optimize the placement of multiple LiDARs for AV while considering the perception performance. Mou \etal \cite{mou2018optimal} formulated a min-max optimization problem for LiDAR placement with a cylinder-based cost function proxy to consider the worst non-detectable inscribed spheres formed by the intersection of different laser beams.
Liu \etal \cite{liu2019should} improved the previous work by using an intuitive volume to surface area (VSR) ratio metric and a black-box heuristic-based optimization method to find the optimal LiDAR placement. These methods assume a uniformly weighted region around the AV to minimize the maximum non-detectable area. They do not explicitly reveal the relation between the LiDAR placement and the perception performance. Recent similar work \cite{ma2021perception} proposed the perception entropy metric for multiple sensor configuration evaluation. Although they use conditional entropy to measure the sensor perception, they rely on empirical assumptions from KITTI dataset in the formulation of perception entropy, which weakens the generalizability to other scenes. We formulate the problem under a full probabilistic framework using a data-driven surrogate metric with a ray-tracing acceleration approach to overcome these limitations.






\section{Methodology}
\label{method}
To avoid the huge effort of data collection and complicated analysis of learning-based detection  for evaluating each multi-LiDAR placement, we introduce the methodology to efficiently evaluate LiDAR sensor placement for perception performance based on information theory. The problem of LiDAR placement  evaluation is formulated  with the definitions of Region of Interest (ROI) and Probabilistic Occupancy Grid (POG), by introducing a probabilistic surrogate metric for fast LiDAR configuration evaluation.

\subsection{Problem Formulation}
\label{problem-formulation}

\begin{figure}[t!]
\begin{center}
 \includegraphics[width=\linewidth,height=0.4\linewidth]{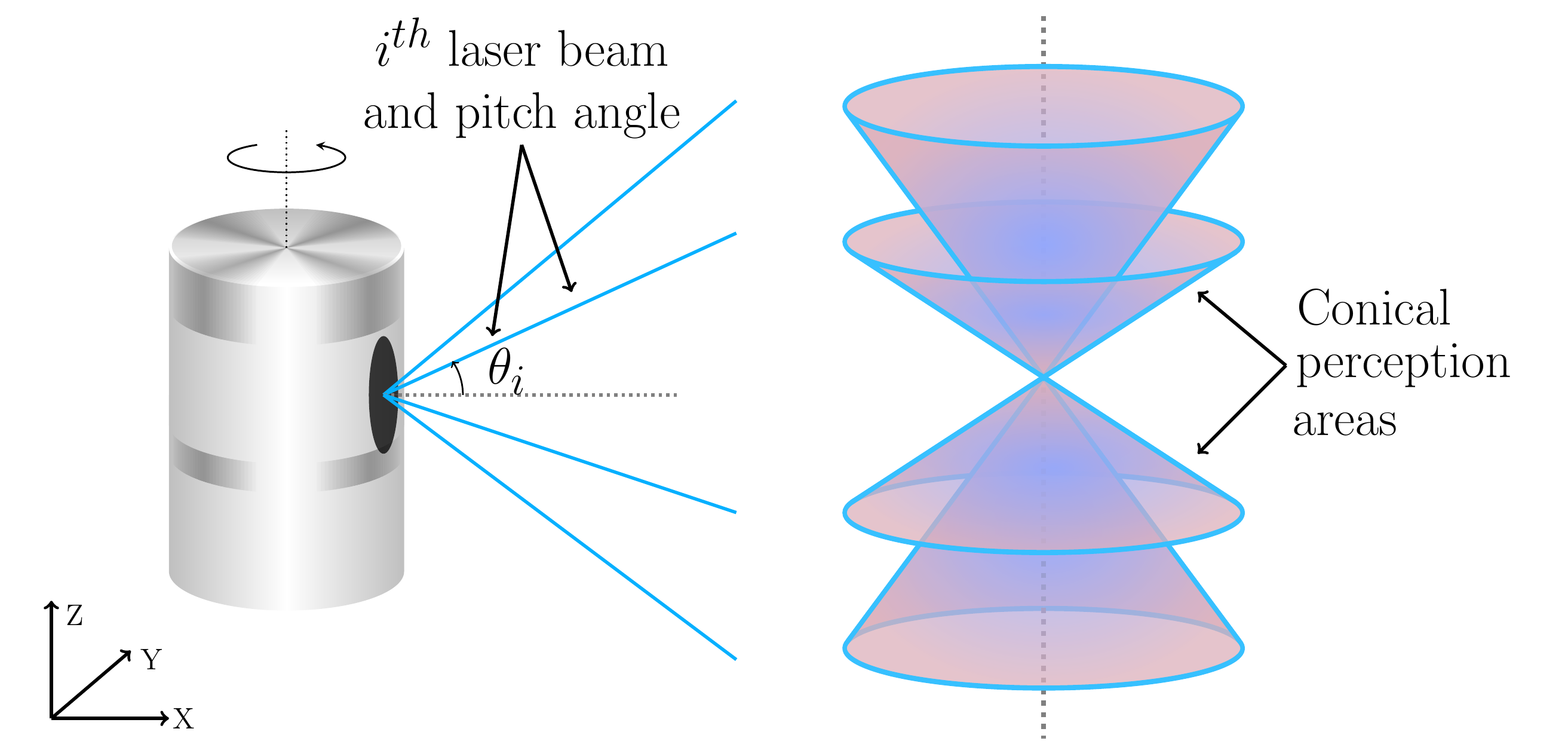}
\end{center}
\vspace*{-5mm}
  \caption{Schematic showing a LiDAR sensor forming perception cones in the ROI to collect point cloud data.}
\label{fig:lidar-schematic}
\end{figure}

We begin this section by defining the LiDAR perception model and the ROI, which form the basis of our LiDAR configuration evaluation problem.
As shown in Figure \ref{fig:lidar-schematic}, we model a LiDAR sensor as a collection of multiple beams. Each beam owns a pitch angle to the $XY$ plane and rotates at a uniform speed along the positive $Z$ axis. As each beam completes one rotation, it forms a conical surface area, and we assume its perception to be all points in this area.  Thus, the total perception of a LiDAR is the union of all these conical areas formed by the rotation of its beams in the ROI, which we will describe next.

\begin{figure}[t!]
\begin{center}
 \includegraphics[width=\linewidth,height=0.65\linewidth]{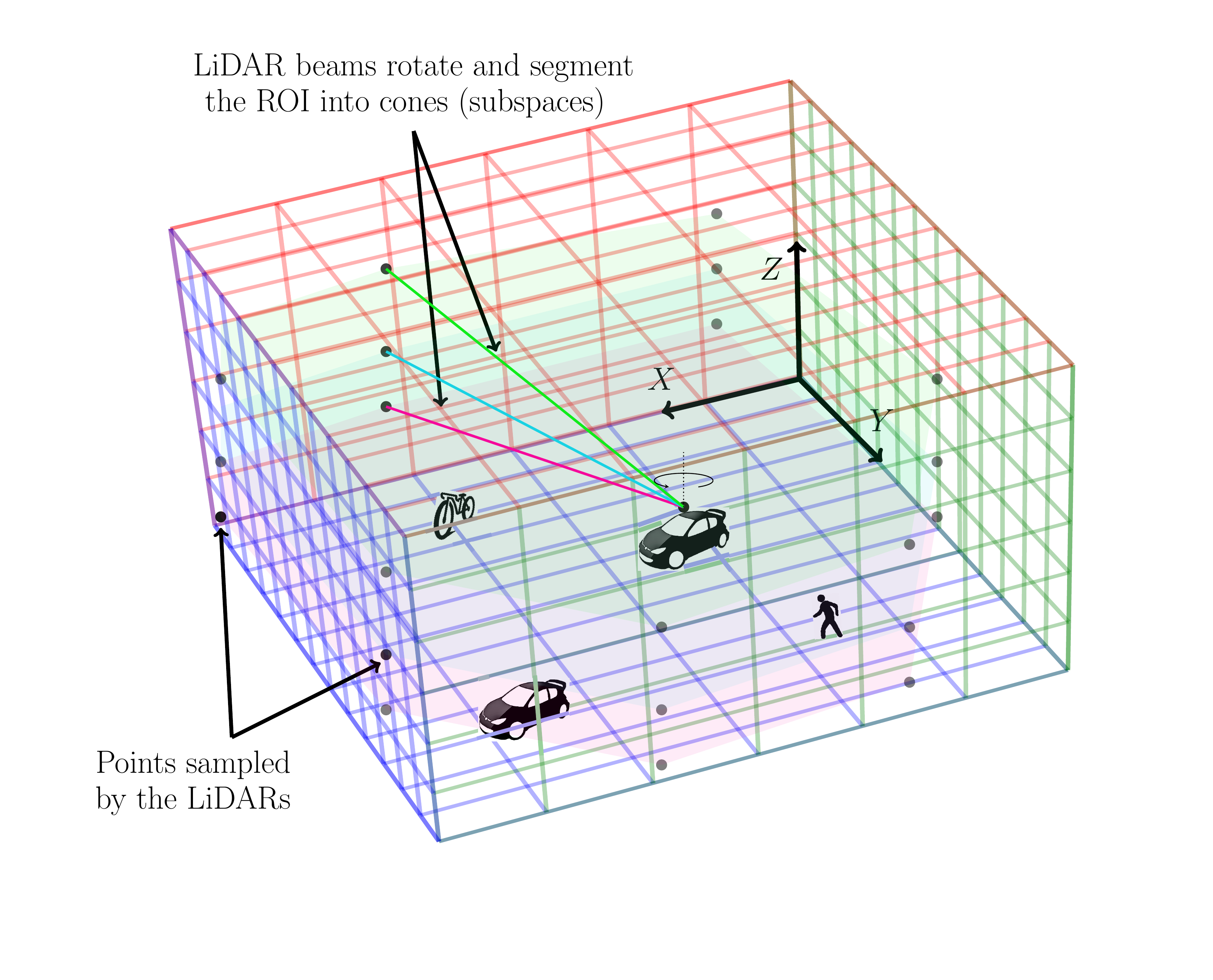}
\end{center}
\vspace*{-12mm}
  \caption{LiDAR sensor mounted on an AV samples object voxels according to 3D bounding boxes from the ROI and generate the POG to evaluate detection performance. }
\label{fig:roi-lidar}
\vspace{-4mm}
\end{figure}

Similar to previous work \cite{liu2019should} and \cite{mou2018optimal}, we define the ROI to be the space where we keep track of objects to be detected. To account for LiDAR's limited range of detection, we denote the cuboid size of the ROI to be $[l, w, h]$ for length $l$ along the x-axis, width $w$ along y-axis and height $h$ along the z-axis in the $XYZ$ coordinate system, as shown in Figure \ref{fig:roi-lidar}. The size of ROI is mainly determined by the distribution of target objects (Car, Truck, Cyclist, \textit{etc.}) in the scenario.
We then discretize the ROI into voxels with a fixed resolution $\delta$ to represent ROI as a collection of voxels,
\begin{equation}
\label{ROI_voxels}
\mathcal{V}=\{v_1, v_2, \dots, v_M\}, M=\frac{l}{\delta}\times\frac{w}{\delta}\times\frac{h}{\delta}
\end{equation}
 The ROI provides us with a fixed perception field around the LiDAR, from which LiDAR beams accumulate most sensing information for the following perception task.

Therefore, for the LiDAR-based 3D object detection task, we only focus on the objects within ROI for each point cloud frame when calculating the perception metrics.
Then, the problem of LiDAR placement evaluation is formulated as comparing the object detection performance given LiDAR configuration candidate using common point cloud-based 3D object detection algorithms and their metrics \cite{geiger2013vision}. However, directly using 3D object detection metrics to evaluate the LiDAR placement in the real world is inaccurate and extremely inefficient, as it is impossible to make all the scenarios and objects identical to collect point cloud data fairly for evaluated LiDAR configuration candidate in the practical application. Besides, it may take days for each evaluation procedure to collect new data and train the detection models based on new LiDAR placement to get the final detection metrics. Therefore, we propose a new surrogate metric to accelerate the LiDAR placement evaluation procedure based on the Probabilistic Occupancy Grid.

\subsection{Probabilistic Occupancy Grid}
Since we consider the 3D object detection in ROI among all the frames, intuitively, the LiDAR configurations covering more objects will perform better in the object detection task. To this end, we propose to model the joint distribution of voxels in ROI as the Probabilistic Occupancy Grid (POG) by estimating the probability of each voxel to be occupied. For each object of interest, like car, truck and cyclist, the POG is defined as the joint probability of occupied voxels by 3D bounding boxes (BBoxes) among all the frames from $M$ voxels in ROI. Similarly, let the LiDAR configuration be a random variable $C$, given LiDAR configuration $C=C_0$, conditional POG can represent the conditional joint distribution of occupied voxels given the specific LiDAR configuration with the assumption of conditional independence.
\begin{align}
\label{pog}
p_{POG} &= p(v_1, v_2,\dots,v_M)  \\
\label{pog_C}
p_{POG \mid C=C_0} & = p(v_1, v_2,\dots,v_M \mid C=C_0)
\end{align}
where $v_i \sim  p_{\mathcal{V}}$. To make the notation compact and easy to read, we denote the occupied voxel random variable   $v_i \mid C=C_0  $  as $v_i^{C_0}$ and denote  the conditional distribution $p_{\mathcal{V} \mid C=C_0}$ as $p_{\mathcal{V} \mid C_0}$, so we have $, v_i^{C_0}  \sim p_{\mathcal{V} \mid C_0}$.
\begin{figure}[t!]
\begin{center}
 \includegraphics[width=\linewidth]{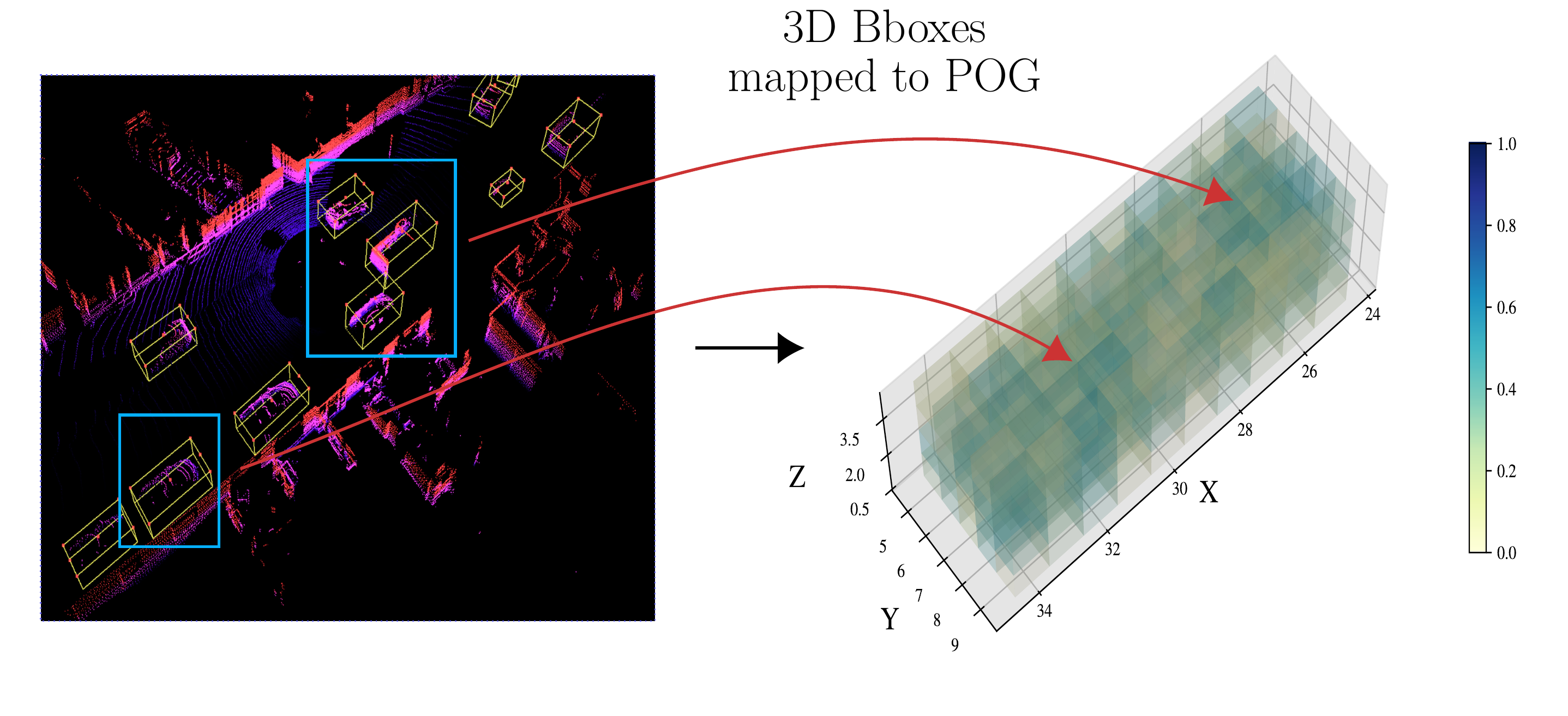}
\end{center}
\vspace*{-8mm}
  \caption{Schematic showing how 3D bounding boxes (Bboxes) are mapped to the ROI to generate a POG. Each cube in the POG has a probability given by Equation {\ref{prob-equation}}.}
\label{fig:pc-to-pog}
\vspace{-13mm}
\end{figure}

To estimate POG from samples, given a dataset of the ground truth of BBoxes $\mathcal{Y}_{T}=\{y_1, y_2,\dots, y_T\}$, where $T$ represents the number of ground truth frames. Each frame $y_t (t\in\{1,...,T\})$ contains $N^{(t)}$ 3D BBoxes of the target object $\{b^{(t)}_1, b^{(t)}_2, \dots, b^{(t)}_{N^{(t)}}\}$ within ROI of the ego-vehicle, and each BBox is parameterized by its center coordinates, size (length, width, height), and yaw orientation. For each voxel coordinate $v_i=(x_i, y_i, z_i) \in \mathcal{V}$, we denote $v_i \in y_t$ if $v_i$ is within any of the bounding boxes $\{b^{(t)}_1, b^{(t)}_2, \dots, b^{(t)}_{N^{(t)}}\}$. Then, for each voxel $v_i \in \mathcal{V}$ from ROI, we estimate its probability to be occupied among all the $T$ frames as,
\begin{align}
\label{prob-equation}
\hat{p}(v_i) = \frac{ \sum_{t=1}^T \mathbbm{1}(v_i \in y_t) }{T},& i = 1,2, \dots, M \\
v_i \in y_t := \exists ~ b^{(t)} \in y_t, & s.t. ~ v_i \in b^{(t)}
\end{align}
where $\mathbbm{1}(\cdot)$ is an indicator function and $M$ is the number of voxels in ROI. The POG can then be estimated by the joint probability of all occupied voxels in ROI. Since the presence of an object in one voxel does not imply presence in other voxels among all the frames in ROI, we could treat these voxels as independent and identically distributed random variables and calculate the joint distribution over all voxels in the set $\mathcal{V}$ as,
\begin{equation}
\label{all_pog_est}
\hat p_{POG} = \hat p(v_1,\dots,v_M) = \prod_{i=1, \hat p(v_i)\neq 0}^{M} \hat p(v_i)
\end{equation}
where $M$ is the total number of non-zero voxels in ROI. One such example of the POG of car is shown in Figure \ref{fig:pc-to-pog}. Note that notations of $\hat p$ with $\hat{hat}$ are the estimated distribution from observed samples, while notations of $ p$ without \textit{hat}  are  the unknown non-random true distribution to be estimated.



For a particular LiDAR configuration $C=C_0$, we employ ray tracing by Bresenham's Line Algorithm \cite{5388473} (denoted as $Bresenham(\mathcal{V},C_0)$) to find all the voxels which intersect with the perception field (ROI) of that LiDAR. We denote the set of these beam-intersected voxels as,
\begin{equation}
\label{bresenham_C}
\mathcal{V}| C_0 = Bresenham(\mathcal{V},C_0) = \{v_1^{C_0}, \dots, v_{M}^{C_0}\}
\end{equation}
where the discretized non-zero voxels from LiDAR beams  give the perception range of LiDAR configuration $C_0$.

The conditional probability of occupied voxels among all frames in ROI given the LiDAR configuration $C_0$ can represent the conditional distribution  from which LiDAR can  get sensing information of the target object. Similar to Equation \ref{all_pog_est},  the conditional POG given LiDAR configuration $C_0$ can be estimated as,
\begin{equation}
\label{all_pog_C_est}
\hat p_{POG \mid C_0} = \hat p(v_1^{C_0}, \dots, v_{M}^{C_0}) = \prod_{i=1, \hat p(v^{C_0}_i)\neq 0}^{M} \hat p(v^{C_0}_i)
\end{equation}

From the perspective for density estimation to find POG, combine Equation \ref{pog}, \ref{pog_C},  \ref{all_pog_est} and \ref{all_pog_C_est}, the true POG and conditional POG given  configuration $C_0$ can be estimated as,
\begin{align}
p_{POG} =  \hat p_{POG}, ~
\label{bridge_pog_C}
p_{POG |C=C_0}  = \hat p_{POG |C_0}
\end{align}


\subsection{Probabilistic Surrogate Metric}
In this section, we derive our surrogate metric based on information theory using the POG $p(v_1,...,v_M)$ and conditional POG $p(v_1,...,v_M\mid C = C_0)$ given LiDAR configuration $C_0$,  evaluating how well the LiDAR placement can sense objects' location in ROI.
To maximize the perception capability, one intuitive way is to reduce the uncertainty of the joint distribution of voxels given specific LiDAR configuration. The total entropy of POG (\textbf{POG Entropy}) is only determined by the scenarios with bounding boxes.
\begin{align}
\label{totla_entropy}
H_{POG} = H(\mathcal{V}) =  \mathbb{E}_{v_i \sim p_{\mathcal{V}}} \sum_{i=1}^{M} \hat H(v_i)
\end{align}
Mutual information (MI) of two random variables can represent uncertainty reduction given the condition. In this case, we  want to evaluate the information gain (\textbf{IG}) of voxels occupied by target objects given the LiDAR configuration candidate. Our insight is that the more IG the LiDAR configuration has, the more information it will contain so the more uncertainty it will reduce. With  $p(C=C_0) = 1$,
\begin{align}
\label{IG}
&IG_{\mathcal{V}, C_0} =  H(\mathcal{V}) - H(\mathcal{V}|C_0) = H(\mathcal{V}) + S_{MIG} \\
\label{SurM_MIG}
&S_{MIG} = -H(\mathcal{V}|C_0) = -\mathbb{E}_{v_i^{C_0} \sim p_{\mathcal{V}| C_0}} \sum_{i=1}^{M} \hat H(v^{C_0}_i)
\end{align}
where we introduce the maximum information gain-based surrogate metric \textbf{S-MIG}, ignoring the total entropy $H(\mathcal{V})$ which is a constant given POG and the same for different LiDAR configurations. The entropy of each voxel can be found below.
$$\hat H(v_i) =  -\hat p(v_i)\log \hat p(v_i) - (1-\hat p(v_i))\log (1-\hat p(v_i)) $$

\section{Experiments}
\label{experiments}
\begin{figure}[t!]
\begin{center}
    \begin{subfigure}[b]{0.235\textwidth}
                \includegraphics[width=\linewidth]{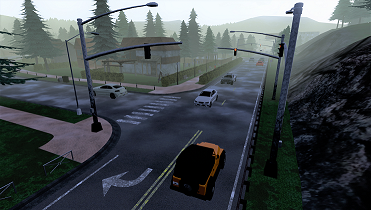}
                \caption{}
                \label{fig:carla-map-4}
    \end{subfigure}
    \begin{subfigure}[b]{0.235\textwidth}
                \includegraphics[width=\linewidth]{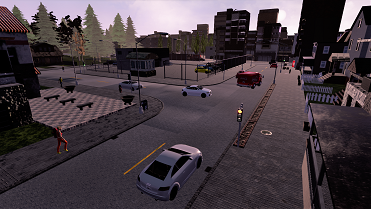}
                \caption{}
                \label{fig:carla-map-3}
    \end{subfigure}%
\end{center}
\vspace{-7mm}
  \caption{Realistic simulation environments in CARLA.}
\label{fig:carla-maps}
\vspace{-5mm}
\end{figure}

This section aims to address two questions: 1) Does LiDAR placement influence the final perception performance? 2) How can we quantify and evaluate the detection performance of different LiDAR configurations using our easy-to-compute probabilistic surrogate metric? To answer these questions, we conduct extensive experiments in a realistic self-driving simulator --- CARLA \cite{dosovitskiy2017carla}. Due to the requirement of fairly evaluating different LiDAR placements with all other environmental factors fixed, such as the ego-vehicle's and surrounding objects' trajectories,
we choose to use realistic simulation scenarios in CARLA (Figure \ref{fig:carla-maps}), instead of a public real-world dataset.



\subsection{Experimental Setup}
 Given CARLA scenarios and the target object, we obtain the POG based on the bounding box labels and calculate the surrogate metric S-MIG for every LiDAR placement. For each evaluated LiDAR configuration candidate,  we first collect point cloud data in CARLA,  then train and test all the object detection models using the collected data. Finally, we correlate the surrogate metric and the 3D detection performance. More details can be found in Appendix \ref{sec:app}.
%

\begin{figure}[h]
\begin{center}
    \begin{subfigure}[b]{0.235\textwidth}
                \includegraphics[width=.98\linewidth]{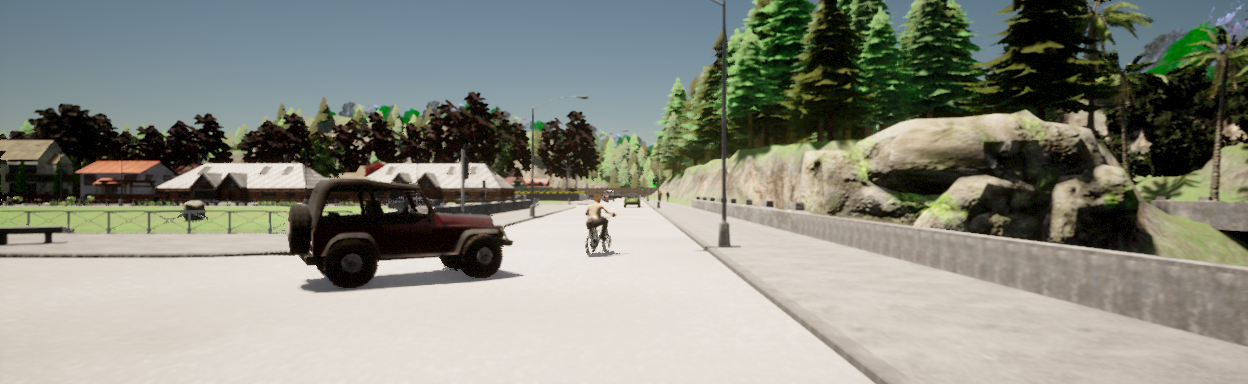}
    \end{subfigure}%
        \begin{subfigure}[b]{0.235\textwidth}
                \includegraphics[width=.98\linewidth]{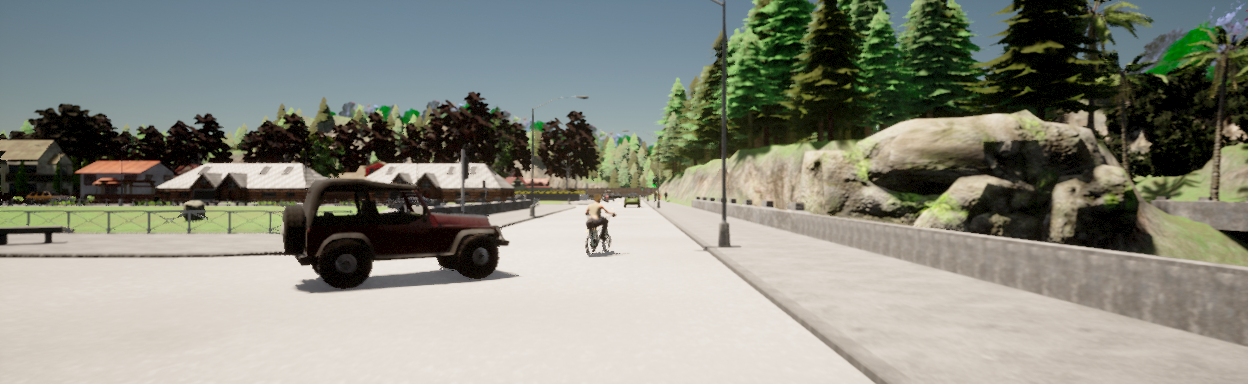}
    \end{subfigure}%

    \begin{subfigure}[b]{0.235\textwidth}
                \includegraphics[width=.98\linewidth,height=.6\linewidth]{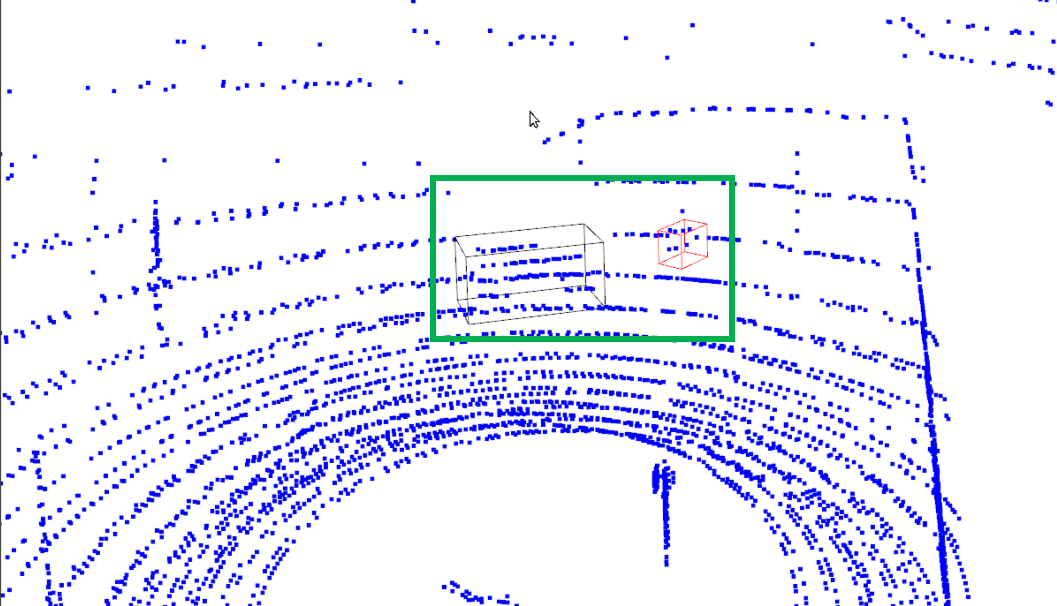}
    \end{subfigure}%
    \begin{subfigure}[b]{0.235\textwidth}
                \includegraphics[width=.98\linewidth,height=.6\linewidth]{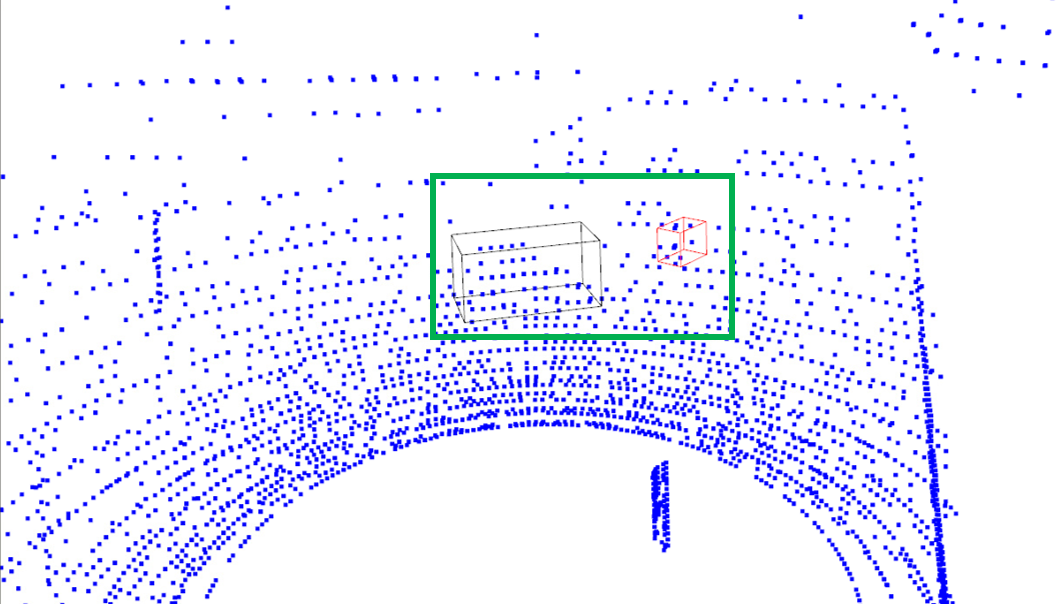}
    \end{subfigure}%

    \begin{subfigure}[b]{0.235\textwidth}
                \includegraphics[width=.98\linewidth,height=.6\linewidth]{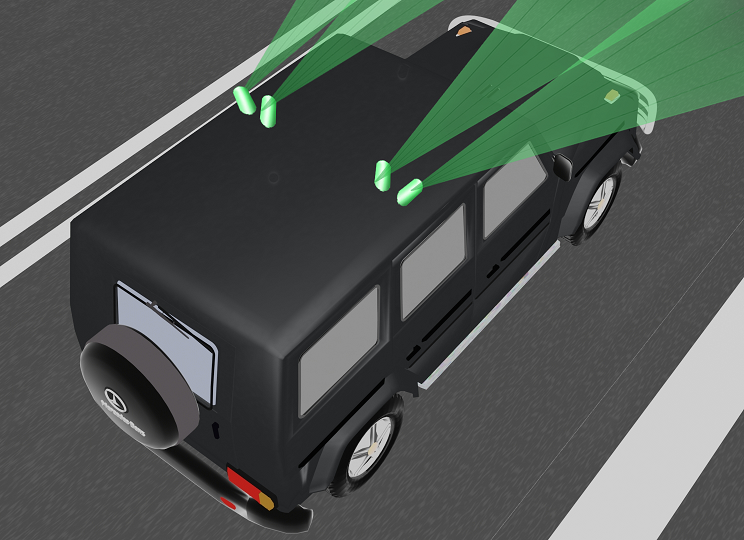}
                \caption{Line-roll}
                \label{fig:baseline-line}
    \end{subfigure}%
    \begin{subfigure}[b]{0.235\textwidth}
                \includegraphics[width=.98\linewidth,height=.6\linewidth]{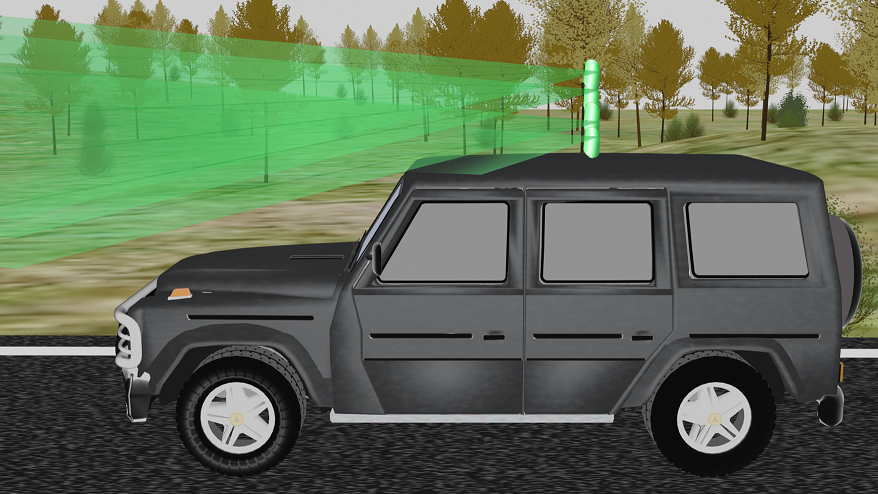}
                \caption{Center}
                \label{fig:baseline-center}
    \end{subfigure}%
\end{center}
\vspace*{-5mm}
  \caption{Point cloud data collected from different LiDAR configurations under the same scenarios with car (in black 3D bounding box) and cyclist (in red 3D bounding  box).
  }
\label{fig:compare-config}
\vspace*{-5mm}
\end{figure}

\textbf{CARLA simulator.} We use CARLA as the simulation platform to collect data and evaluate our method. CARLA is a high-definition open-source simulator for autonomous driving research \cite{chen2020learning,chen2019model} that offers flexible scenario setups and sensor configurations. CARLA provides realistic rendering and physics simulation based on Unreal Engine 4. Furthermore, the LiDAR sensor could generate accurate point cloud data and could be easily configured with customized placement.
We use CARLA's inbuilt \texttt{ScenarioRunner} module to simulate  realistic traffic scenarios, only changing  LiDAR configurations for all the experiments. We collect data in \texttt{Town 1, 3, 4, 6} and each town contains 8 manually recorded routes. For \texttt{Sparse}, \texttt{Medium} and \texttt{Dense} scenarios, we spawn 40, 800, 120 Vehicles (including cars, motorcycles, trucks, cyclists, etc.) in each town and the default density  is \texttt{Medium}.



\textbf{Data format.}
The collected dataset for each LiDAR configuration contains about 45000 point cloud frames with 3D bound boxes of normal-size vehicles as Car (including Sedan, Pickup Truck, SUV, etc.) and other abnormal-size vehicles as Van and Cyclist (including Box Truck, Cyclist, Motorcycles, etc.). We use all collected bounding boxes to calculate POG for each object type and estimate the joint distribution of voxels in ROI. We use about 10\% fixed frames as the test set and the remaining data as the training set in all the experiments. The point clouds from different LiDARs are aggregated and transformed to the reference frame of the ego-vehicle   for computational convenience. We make the data format consistent with KITTI 3D object detection benchmark
\cite{geiger2013vision}. To calculate ROI and POG efficiently for each target object, we customize the size of ROI to be $l=80m, w=40m, h=4m$ for CARLA scenario. Following the KITTI format, we use the half front view of point cloud range (40m in length)  for POG and surrogate metric, and the 3D detection model training and testing.

\textbf{3D object detection algorithms and metrics.} To fairly compare the object detection performance of different LiDAR configurations, we adopt the recent open-source LiDAR-based 3D object detection framework \textit{OpenPCDet} \cite{openpcdet2020} and fine-tune the KITTI pretrained models of multiple representative 3D detection algorithms. For voxel-based methods, we use one-stage SECOND \cite{yan2018second} and two-stage Voxel RCNN  \cite{deng2020voxel}. Also, we use PointRCNN \cite{Shi_2019_CVPR} as point-based method and PV-RCNN  \cite{shi2020pv} as the integration of voxel-based and point-based method. Moreover, we also compare the models with 3D Intersection-over-Union (IoU) loss for SECOND \cite{yan2018second} and PointRCNN \cite{Shi_2019_CVPR}.
For detection metrics, we adopt the strictest metrics of Bird-Eye-View (BEV) and 3D detection with average precision across 40\% recall under IoU thresholds of 0.70 for Car and 0.50 for Van and Cyclist \cite{geiger2013vision, openpcdet2020}.
We follow the default training hyperparameters and fine-tune  all the KITTI pretrained models for 10 epochs on our training set for the fair comparison between different LiDAR placements.

\begin{table*}[]
\centering
\resizebox{0.9\textwidth}{!}{
\begin{tabular}{|c||c|c|c|c||c|c|c|c|}
\hline
\multirow{2}{*}{Models}          & \multicolumn{4}{c||}{ Car-3D  (AP\_R40@0.70)}                                          & \multicolumn{4}{c|}{Car-BEV (AP\_R40@0.70)}                                 \\ \cline{2-9}
                                    & Center & Line & Pyramid & Trapezoid   & Center & Line &  Pyramid & Trapezoid \\ \hline
PV-RCNN  \cite{shi2020pv}
     &     52.15 &	55.50 &	\textbf{57.44} &	57.37
 &	62.64&	65.42&	\textbf{65.81}&	65.54
                    \\ \hline
Voxel RCNN  \cite{deng2020voxel}
&47.59	&50.85&	\textbf{52.61}&	50.95
   & 57.72&	60.84&	\textbf{63.65}&	61.52
                                                 \\ \hline
PointRCNN \cite{Shi_2019_CVPR}
&	38.46 &	\textbf{48.25}&	44.28&	47.50
& 51.41&	\textbf{59.14}&	56.97&	59.08
                 \\ \hline
PointRCNN-IoU \cite{Shi_2019_CVPR}
&	38.44 &	46.86&	45.34&	\textbf{47.32}
&	50.56&	58.86&	56.67&	\textbf{59.04}
                         \\ \hline
SECOND \cite{yan2018second}
& 42.89&	47.13&	46.89&	\textbf{47.53}
& 56.65&	59.44&	\textbf{61.75}&	59.97
                                \\ \hline
SECOND-IoU  \cite{yan2018second}
&	44.41&	48.60&	\textbf{51.03}&	49.10
 &	56.98&	59.96&	\textbf{62.31}&	60.28
                                 \\ \hline
\hline
\multirow{2}{*}{Models}          & \multicolumn{4}{c||}{ Van and Cyclist-3D  (AP\_R40@0.50)}                                          & \multicolumn{4}{c|}{Van and Cyclist-BEV (AP\_R40@0.50)}                                 \\ \cline{2-9} &
Center & Line & Pyramid & Trapezoid &
Center & Line &  Pyramid & Trapezoid \\ \hline
PV-RCNN  \cite{shi2020pv}
&	40.09 &	39.11 &	40.26 &	\textbf{42.85}
 &	41.18&	41.45&	43.09&	\textbf{44.79}
                   \\ \hline
Voxel RCNN  \cite{deng2020voxel}
&	33.76&	31.60&	33.39&	\textbf{33.91}
   &	35.40&	33.24&	35.11&	\textbf{35.54}
                   \\ \hline
PointRCNN \cite{Shi_2019_CVPR}
&	\textbf{31.43}&	28.00&	27.10&	30.91
   &	\textbf{33.86}&	30.68&	30.49&	33.75
                      \\ \hline
PointRCNN-IoU \cite{Shi_2019_CVPR}
&	\textbf{30.75}&	27.64&	26.19&	30.54
 &	34.04&	30.71&	29.30&	\textbf{34.70}
                  \\ \hline
SECOND \cite{yan2018second}
&	\textbf{36.13}&	32.36&	35.32&	35.95
 &	38.19&	30.68&	\textbf{40.84}&	40.01
                     \\ \hline
SECOND-IoU  \cite{yan2018second}
&	35.61&	33.12&	34.95&	\textbf{36.51}
&	38.99&	36.60&	38.31&	\textbf{40.14}
                        \\ \hline
\end{tabular}}
\caption{Comparison of object detection performance under various LiDAR configurations  using different algorithms.
}
\label{tab:expt1}
\vspace*{-5mm}
\end{table*}

\textbf{Different LiDAR Placements} We evaluate several LiDAR placements as baselines to show the influence on object detection performance. We mainly consider the placement problem of 4 LiDARs and each LiDAR has 20Hz rotation frequency and 90,000 points per frame from the 16 beams.
In our experiments, we set beams to be equally distributed in the vertical FOV $[-25.0, 5.0]$ degrees.
\textit{Trapezoid} LiDAR configuration is simplified from  Toyota's self-driving cars with 4 LiDARs in the parallel front and back (Figure~\ref{fig:toyota}). \textit{Pyramid} placement is motivated by Cruise and Pony AI (Figure~\ref{fig:cruise}, \ref{fig:ponyai}) including 1 front LiDAR and 3 back ones with a higher one in the middle.
 \textit{Center} placement is achieved by vertically stacking four LiDARs together at the center of the roof, which is inspired by Argo AI's autonomous vehicle (Figure~\ref{fig:argoai}). \textit{Line} is motivated by Ford (Figure~\ref{fig:ford}) and 4 LiDARs are placed in a horizontal line symmetrically.  The idea of \textit{Square} comes from Zoox's self-driving cars (Figure~\ref{fig:zoox}), which places the 4 LiDARs on the 4 roof corners.
Furthermore, we investigate the influence of roll rotation on sided LiDARs (\textit{Line-roll}, \textit{Pyramid-roll}).
The visualization of some baseline placements is shown in Figure~\ref{fig:compare-config}
and more details of LiDAR placement are presented in the Figure \ref{fig:lidar-placement}.

\subsection{Experimental Results and Analysis}

To answer the questions raised in Section \ref{experiments}, we demonstrate our evaluation results first to  show the influence of different LiDAR placements on 3D object detection, then analyze the relation between detection performance and our surrogate metric in detail. Note that we limit the number of input points per frame to make the detection challenging enough for multi-LiDAR configuration, resulting in lower detection metric values than the original KITTI benchmark \cite{geiger2013vision}.  See Appendix \ref{app:more_ressults} for more details.

\begin{table}[]
\centering
\resizebox{0.45\textwidth}{!}{
\begin{tabular}{|c|c|c|c|}
\hline
3D    & PV-RCNN\cite{shi2020pv}  &  PointRCNN\cite{Shi_2019_CVPR}  & \makecell[c]{S-MIG\\ $(10^3)$}  \\ \hline
Line  &   55.50
      &     48.25
      &  -5.02      \\ \hline
Line-roll &  54.53
&    45.12
      &    -6.05       \\ \hline
Pyramid &    57.44
    &   44.28
       &   -5.64       \\ \hline
Pyramid-roll &   53.91
     &    36.91
      &    -6.69      \\ \hline
\hline
BEV    & PV-RCNN\cite{shi2020pv}   & PointRCNN\cite{Shi_2019_CVPR}  & \makecell[c]{S-MIG\\ $(10^3)$} \\ \hline
Line  &   65.42
      &     59.14
      &     -5.02     \\ \hline
Line-roll &   63.17
     &    56.54
      &   -6.05       \\ \hline
Pyramid &  65.81
       &    56.97
      &    -5.64     \\ \hline
Pyramid-roll &     63.89
    &     50.74
      &   -6.69     \\ \hline
\end{tabular}}
\caption{Influence of roll rotation of sided LiDARs (as Figure \ref{fig:baseline-line} shows) on Car detection performance.
}
\label{tab:expt4}
\vspace*{-5mm}
\end{table}

\begin{figure*}[t]
  \centering
  \begin{subfigure}[b]{0.45\textwidth}
  \includegraphics[width=\linewidth]{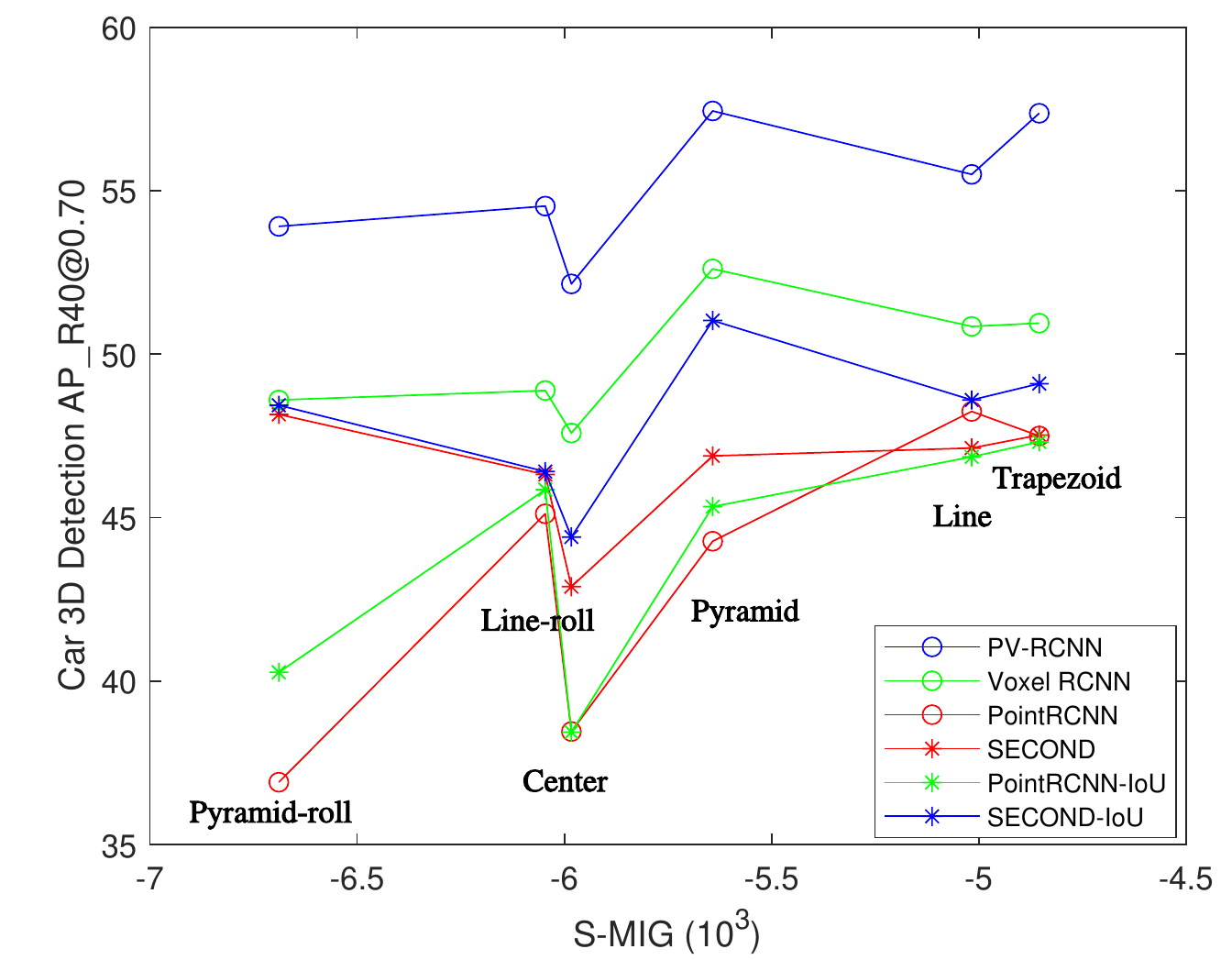}
    \end{subfigure}%
    \begin{subfigure}[b]{0.45\textwidth}
  \includegraphics[width=\linewidth]{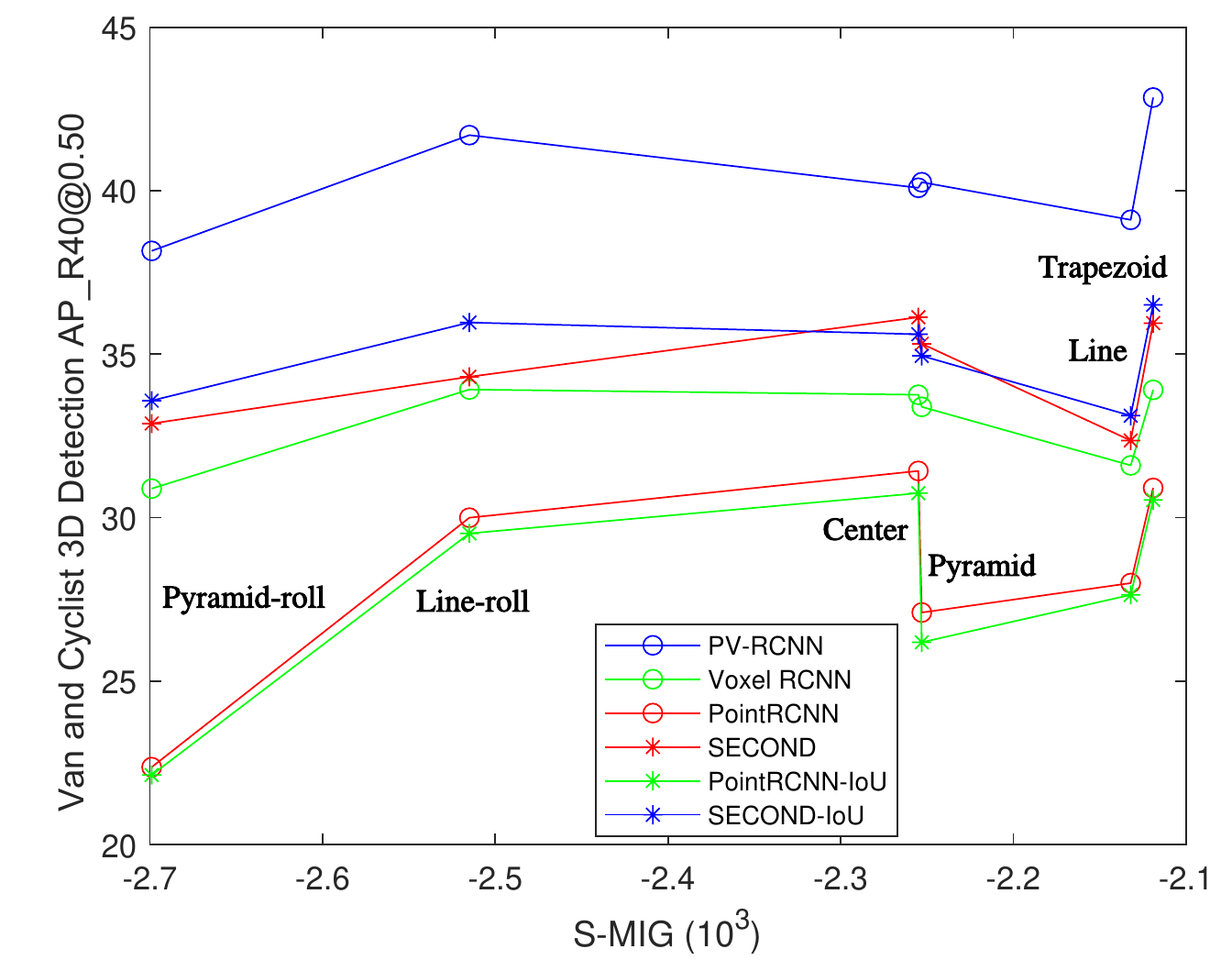}
    \end{subfigure}%

    \begin{subfigure}[b]{0.45\textwidth}
  \includegraphics[width=\linewidth]{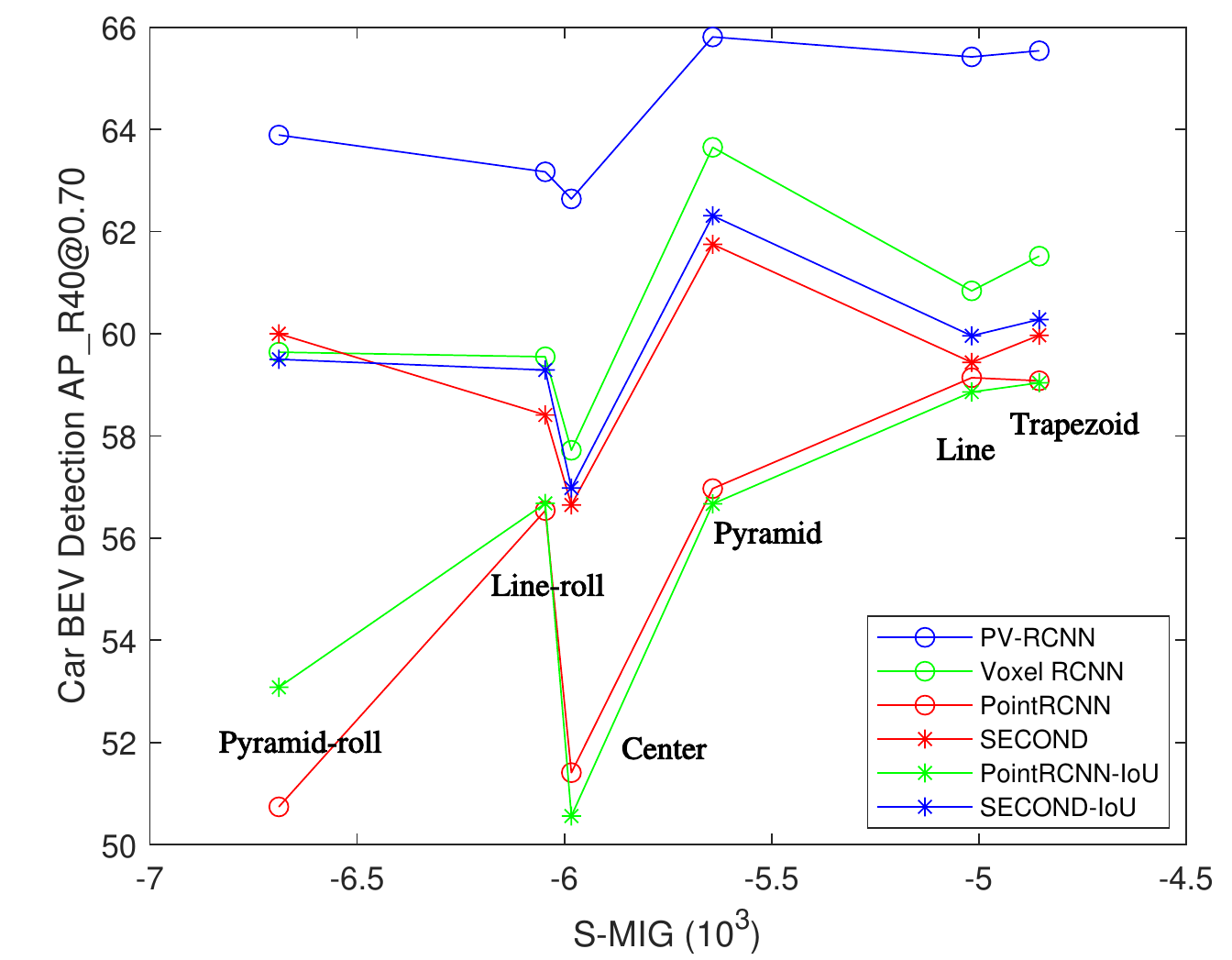}
    \end{subfigure}%
    \begin{subfigure}[b]{0.45\textwidth}
  \includegraphics[width=\linewidth]{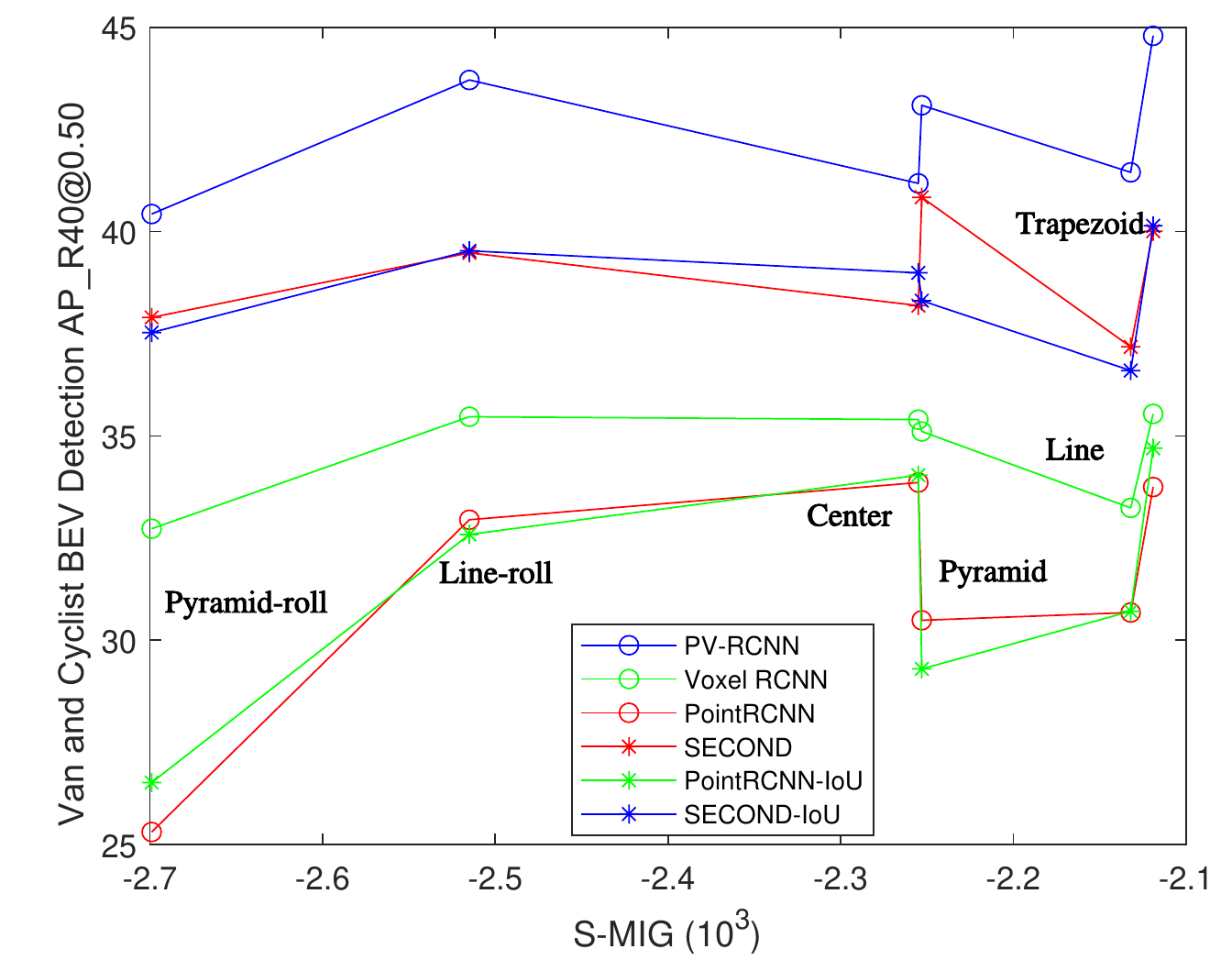}
    \end{subfigure}%
    \vspace*{-3mm}
  \caption{The relationship between Car, Van and Cyclist detection performance  and surrogate metric of different LiDAR placements.}
  \label{fig:expt2}
   \vspace*{-5mm}
\end{figure*}

\textbf{LiDAR placement influence on 3D object detection.} From Table \ref{tab:expt1} and Figure \ref{fig:expt2}, we present the 3D detection (3D IoU) and bird-eye-view (BEV) performance of representative voxel-based and point-based detection algorithms using point cloud collected with different LiDAR configurations. Note that we use the most rigorous detection metrics in our experiments from \cite{openpcdet2020}. It can be seen that different LiDAR placements clearly influence the detection performance for all the algorithms, varying 10\% at most. Moreover, for different target objects, the influence of LiDAR placement is quite different as well. \textit{Pyrimid} configuration perform better on Car detection for most algorithms. In contrast, most models trained with data collected with \textit{Trapezoid} placement has  higher detection precision for Van and Cyclist object. The reason lies in different POG and point cloud distribution between Car, Van and Cyclist, so the most suitable placement for different target objects is different.

\textbf{Relation between detection performance and surrogate metric.} Now we will show why different LiDAR placements will affect the detection performance using surrogate metric S-MIG. As shown in Figure \ref{fig:expt2}, we illustrate the relation between Car, Van and Cyclist detection performance and S-MIG with all the LiDAR placements using both 3D and BEV average precision metrics. As S-MIG increases, the 3D and BEV detection metrics under all algorithms generally go up for Car, Van and Cyclist detection. The fluctuation in the plots comes from some noise and outliers in data collection and model training, but the increasing trend of detection performance with respect to S-MIG is  revealed, which explains the influence of LiDAR placement. More specifically, S-MIG is smoother in reflecting detection of small objects like cyclists or extremely large objects like box trucks.
From Figure \ref{fig:compare-config} we can find that  point cloud collected through \textit{Center} is more uniformly distributed so that cyclists get more points  compared to \textit{Line-roll}, while points on large objects may get saturated so the evaluation with S-MIG may be affected.



\begin{table}[]
\centering
\resizebox{0.45\textwidth}{!}{
\begin{tabular}{|c|c|c|c|}
\hline
Density & Dense  & Medium  & Sparse \\ \hline
PV-RCNN   \cite{shi2020pv}                 &  57.98
     &  56.88
      &  54.48
      \\ \hline
PointRCNN   \cite{Shi_2019_CVPR}                &  48.13
     &  45.46
      &  42.30
      \\ \hline
SECOND \cite{yan2018second}                  &  48.19
     &  47.57
     &  43.96
      \\ \hline
Voxel RCNN  \cite{deng2020voxel}                &  52.15
      &  51.14
    &  48.15
     \\ \hline
\hline
S-MIG $(10^3)$                 &   -10.68    &  -7.63       &   -6.27     \\ \hline
 $H_{POG}$(\ref{totla_entropy}) $(10^3)$                 &   618.40    &  496.70      &   416.96     \\ \hline
IG (\ref{IG}) $(10^3)$                        &   607.072    &  489.06      &  410.70      \\ \hline
\end{tabular}
}
\caption{Influence of scenarios with different densities of Car on \textit{Square} 3D detection.
}
\vspace*{-5mm}
\label{tab:expt5}
\end{table}

\subsection{Ablation Study and Application Analysis}
\label{sec:ablation}
In this section, we further investigate the influence on placement-detection correlation from some key factors in self-driving, giving  examples to evaluate the LiDAR configurations using S-MIG towards  potential applications.




\textbf{Roll angles of sided LiDARs.} Orientation of the sided LiDARs are very common, so we investigate the influence of such placement variance in Table \ref{tab:expt4}.  It can be found that the placements of \textit{Line-roll} and \textit{Pyramid-roll} with roll angles  have worse detection performance as their S-MIG values are lower compared to \textit{Line}  and  \textit{Pyramid} respectively. The finding is consistent with detection metrics of both 3D and bird-eye-view, which further validates the effectiveness of our surrogate  metric and gives an example to fast evaluate the LiDAR placement for object detection.

\textbf{Density of surrounding road objects.}
We investigate how scenarios with different object densities affect detection performance and surrogate metric under \textit{Square} placement. Since under scenarios of different object densities, the POG is different and S-MIG cannot be directly used. So, the original information gain IG (\ref{IG}) is adopted to involve different POG entropy $H_{POG}$(\ref{totla_entropy}). From Table \ref{tab:expt5} we can see that the performance gets better when the density of cars increases because the information gain IG gets larger, which shows that our surrogate metric generalizes well to evaluate detection performance under different scenarios.

\textbf{Potential application analysis.}
Based on the correlation of our surrogate metric and detection performance, the evaluation of different LiDAR placement can be largely accelerated without following the time-consuming procedure: LiDAR installation $\rightarrow$ massive data collection $\rightarrow$ model training $\rightarrow$ evaluation of perception performance. Instead, we only needs a 3D bounding box dataset of object of interest to generate the POG and evaluate the LiDAR placement, which is fast and economical, and can also be customized in different deployment scenarios.


Moreover, using the proposed surrogate metric, it is easy to optimize the LiDAR placement given the number of LiDARs and their beams under specific scenarios with the object of interest.  It can maximize the efficacy of the LiDAR sensor and relevant to the AV research community and industry since the current LiDAR placements are more or less intuition driven.
Besides, combining with some recent active perception work  \cite{bartels2019agile,ancha2020active,raaj2021exploiting,ancha2021active}, the proposed surrogate metric in this paper could serve as a guidance for those active sensors to focus on important areas around the AV for different scenarios and target objects.

\section{Conclusion}
\label{conclusion}

This paper investigates the interplay between LiDAR placement and 3D detection performance. We proposed a novel efficient framework to evaluate multi-LiDAR placement and configuration for AVs.
We proposed a data-driven surrogate metric that characterizes the information gain in the conical perception areas, which helps accelerate the LiDAR placement evaluation procedure for LiDAR-based detection performance. Finally, we conducted extensive experiments in CARLA, validating the correlation between perception performance and surrogate metric of LiDAR configuration through representative 3D object detection algorithms.
Research in this paper sets a precedent for future work to optimize the placement of multiple LiDARs and co-design the sensor placement and perception algorithms.


{\small
\bibliographystyle{ieee_fullname}
\bibliography{egbib}
}
\clearpage
\appendix
\section{Appendix - Details of Experimental Settings}
\label{sec:app}
\subsection{Data Collection Details in CARLA}
\begin{figure}[t]
  \centering
  \begin{subfigure}[b]{0.2\textwidth}
  \includegraphics[width=0.95\linewidth]{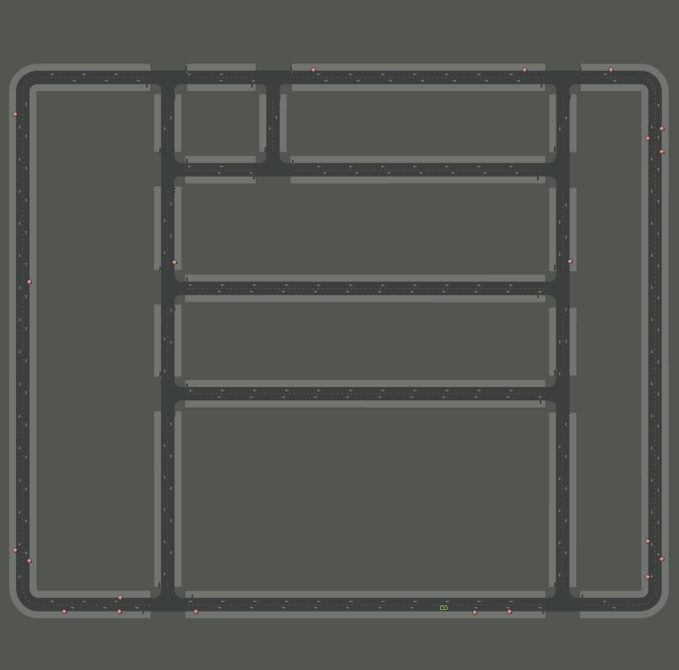}
                \caption{Town 1}
    \end{subfigure}%
  \begin{subfigure}[b]{0.2\textwidth}
  \includegraphics[width=0.95\linewidth]{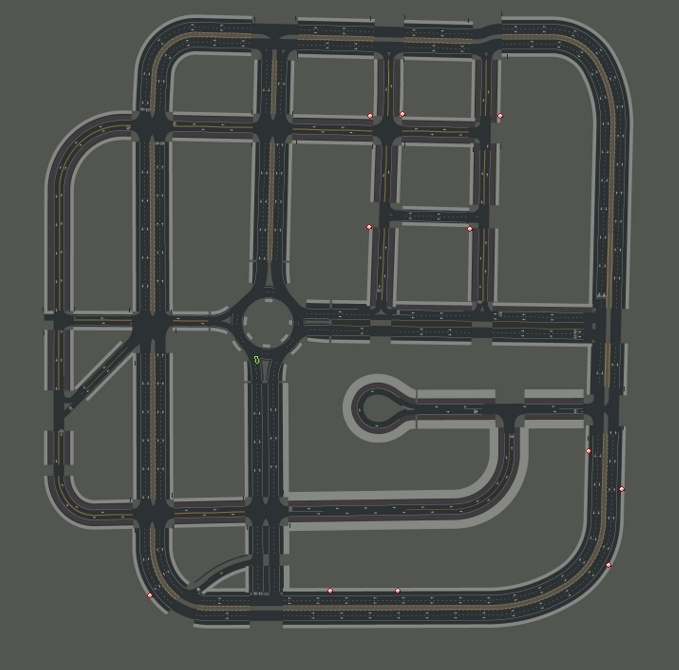}
                \caption{Town 3}
    \end{subfigure}%

      \begin{subfigure}[b]{0.2\textwidth}
  \includegraphics[width=0.95\linewidth]{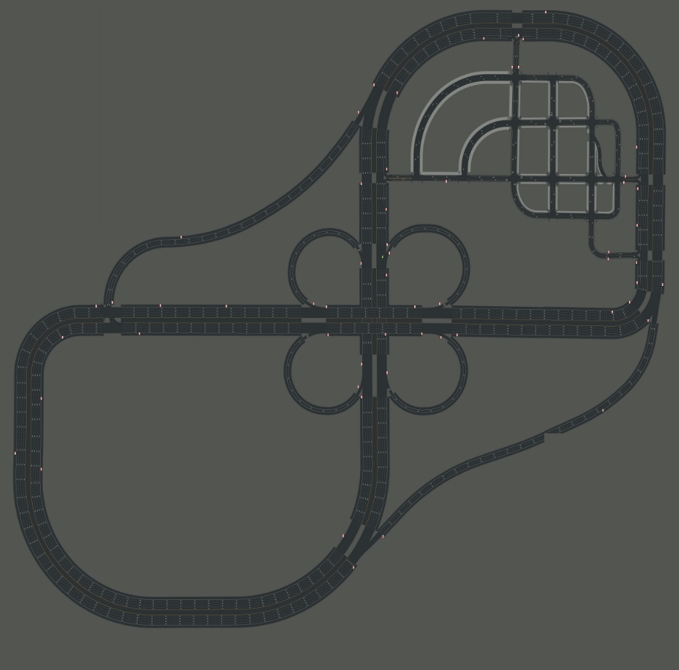}
                \caption{Town 4}
    \end{subfigure}%
      \begin{subfigure}[b]{0.2\textwidth}
  \includegraphics[width=0.95\linewidth]{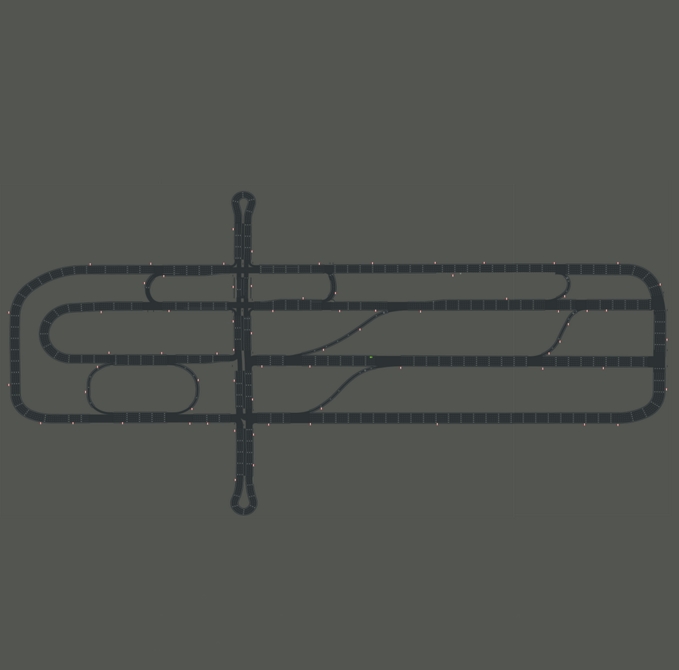}
                \caption{Town 6}
    \end{subfigure}%
  \caption{The four diverse town maps we used to collect data and conduct experiments in CARLA v0.9.10.}
  \label{fig:maps}
   \vspace*{-5mm}
\end{figure}

We choose four different Towns for data collection from CARLA v0.9.10, which are shown in Figure \ref{fig:maps} and the number of frames in each town is about 11000 with 8 different routes covering all the main roads. To split objects, since CARLA itself does not separate Car, Van and Cyclist from Vehicles, we spawn all types of Vehicles and manually denote Van and Cyclist with the actor IDs of \textit{carlamotors.carlacola}, \textit{harley-davidson.low\_rider}, \textit{diamondback.century}, \textit{yamaha.yzf}, \textit{bh.crossbike}, \textit{kawasaki.ninja}  and \textit{gazelle.omafiets} while denoting the remaining as Cars. Note the box truck for \textit{carlamotors.carlacola} is with size over $5.2m \times 2.4m \times 2.6m$, which is the only too large van and categorized into Van and Cyclist, occupying about one tenth of frames in the abnormal-size class.

When collecting point cloud, we choose the frequency of simulation frame to be 20Hz for synchronization and run CARLA on two NVIDIA GeForce RTX 3090 GPUs with RAM 120G in a Ubuntu 18.04 docker container.

\subsection{LiDAR Placement Details}
\begin{figure}[t]
\begin{center}
 \includegraphics[width=0.87\linewidth]{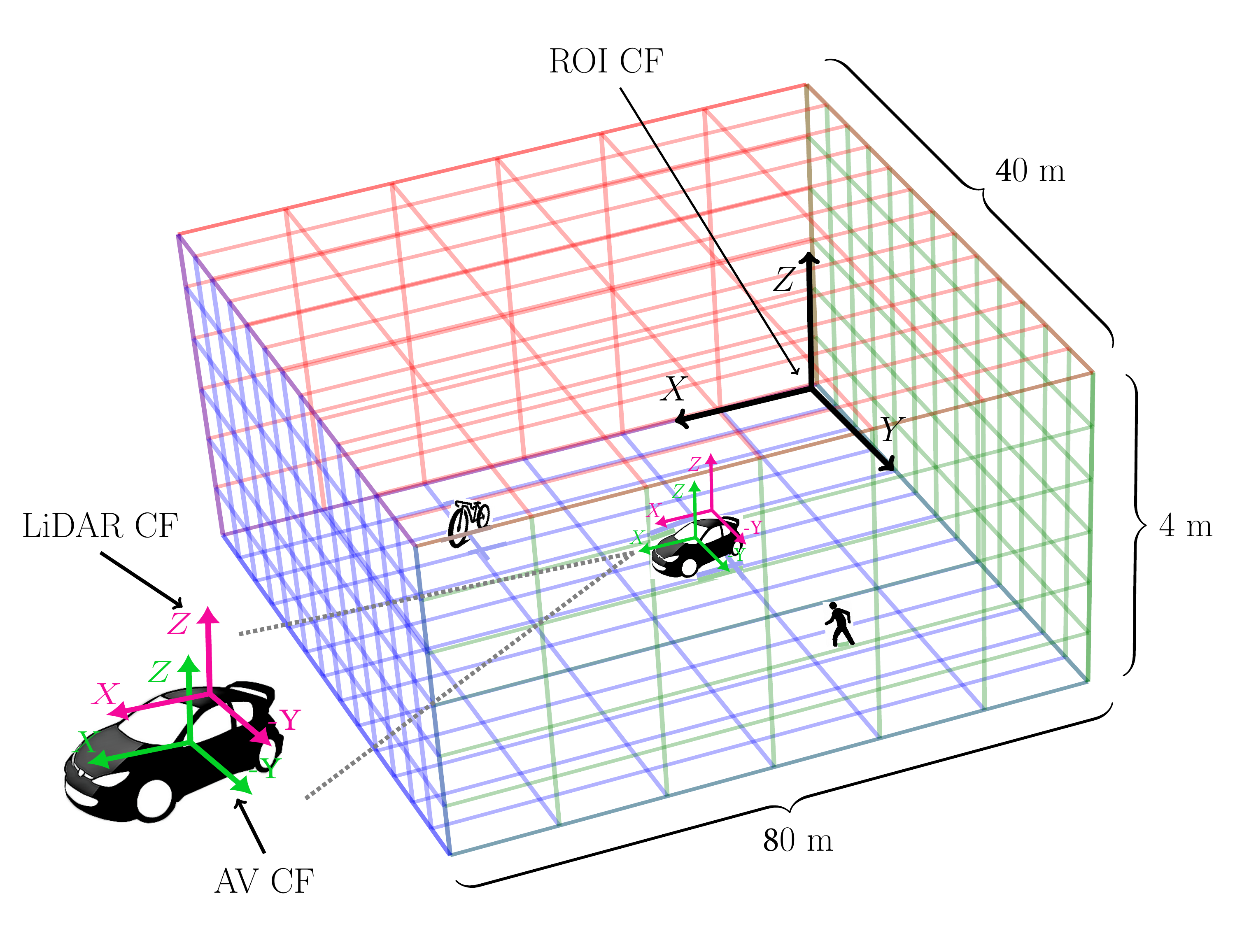}
\end{center}
\vspace*{-5mm}
   \caption{Coordinate frames of ROI, the ego-vehicle, and the LiDAR. Figure not to scale.}
\label{fig:all-frames}
\vspace*{-4mm}
\end{figure}

\begin{figure*}[t]
  \centering
    \begin{subfigure}[b]{0.23\textwidth}
  \includegraphics[width=\linewidth,height=.58\linewidth]{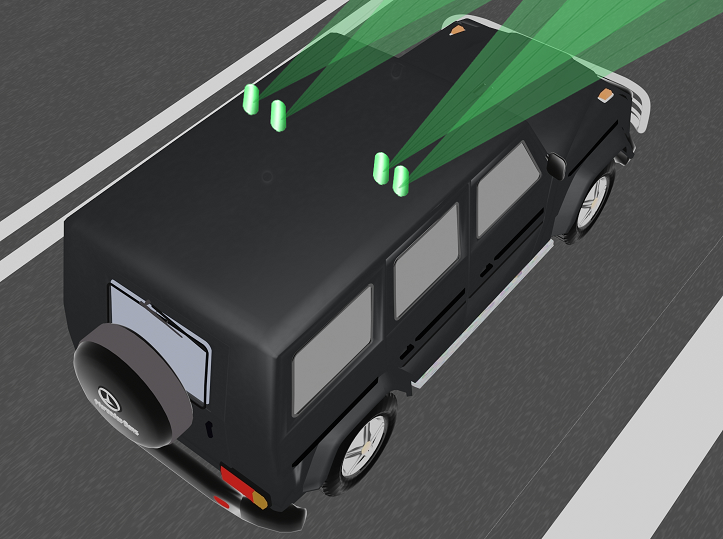}
                \caption{Line}
    \end{subfigure}%
    \begin{subfigure}[b]{0.23\textwidth}
  \includegraphics[width=\linewidth,height=.58\linewidth]{images/baseline-center.png}
                \caption{Center}
    \end{subfigure}%
    \begin{subfigure}[b]{0.23\textwidth}
  \includegraphics[width=\linewidth,height=.58\linewidth]{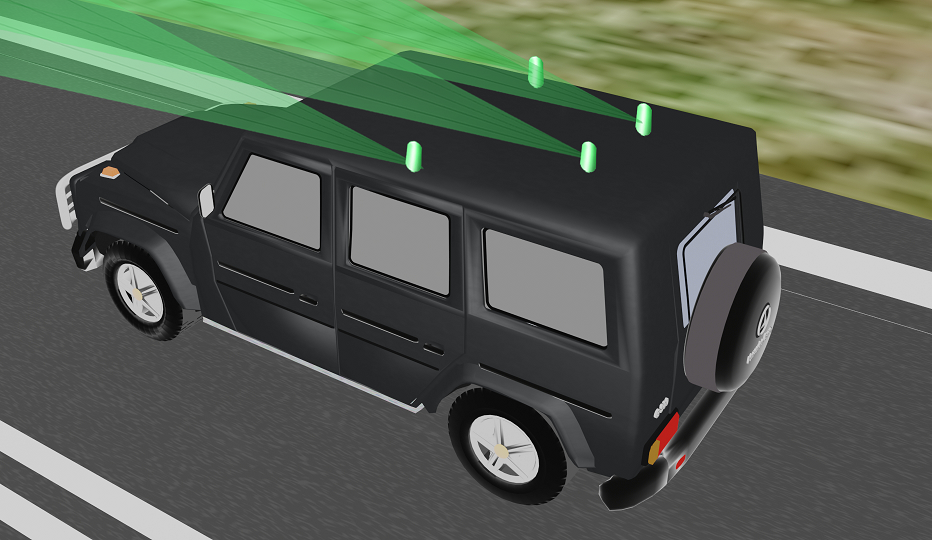}
                \caption{Trapezoid}
    \end{subfigure}%
    \begin{subfigure}[b]{0.23\textwidth}
  \includegraphics[width=\linewidth,height=.58\linewidth]{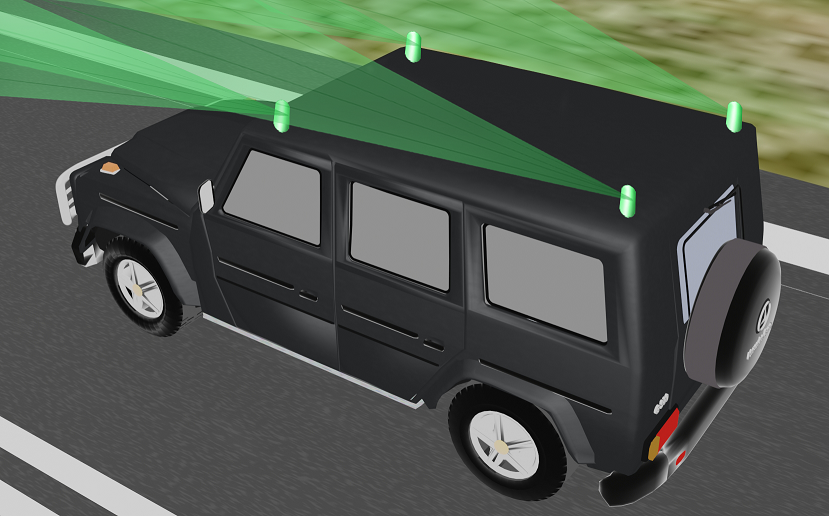}
                \caption{Square}
    \end{subfigure}%

    \begin{subfigure}[b]{0.23\textwidth}
  \includegraphics[width=\linewidth,height=.58\linewidth]{images/baseline-line.png}
                \caption{Line-roll}
    \end{subfigure}%
    \begin{subfigure}[b]{0.23\textwidth}
  \includegraphics[width=\linewidth]{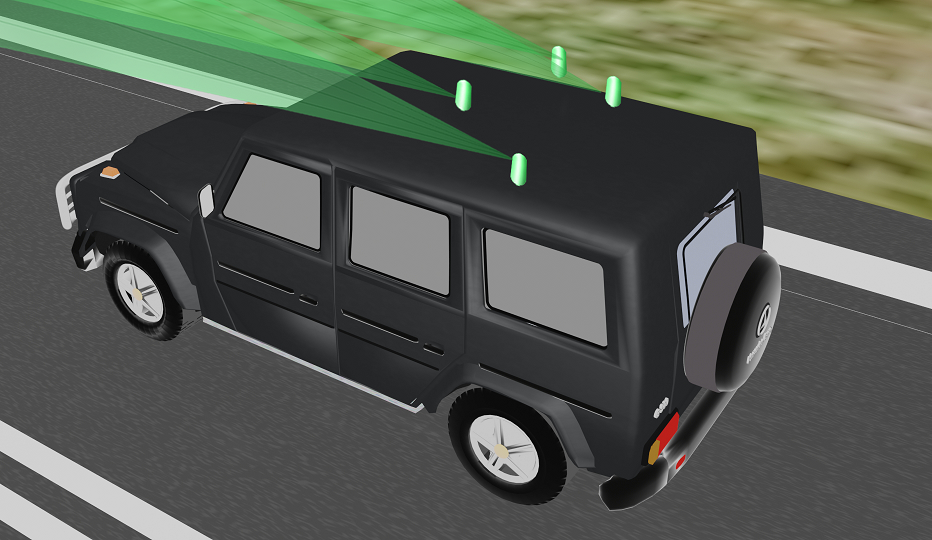}
                \caption{Pyramid}
    \end{subfigure}%
    \begin{subfigure}[b]{0.23\textwidth}
  \includegraphics[width=\linewidth]{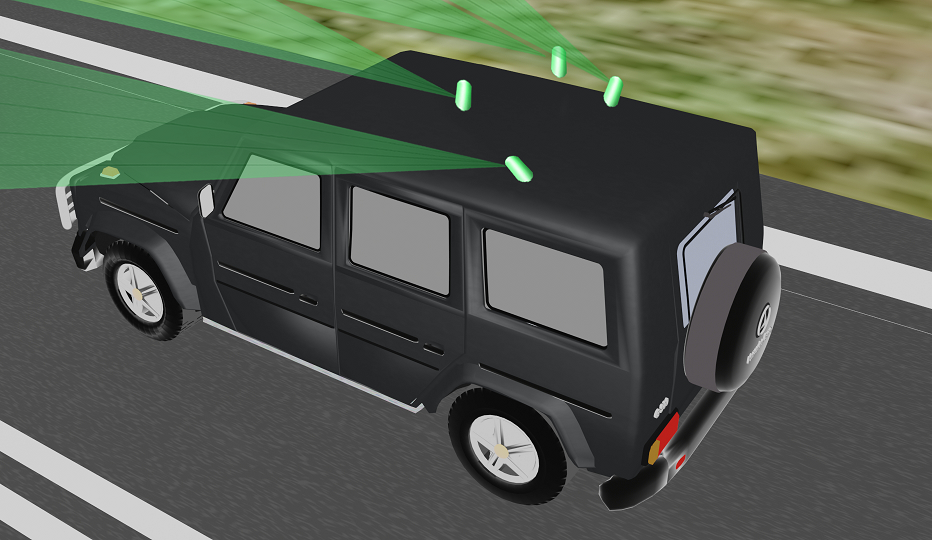}
                \caption{Pyramid-roll}
    \end{subfigure}%
    \begin{subfigure}[b]{0.23\textwidth}
  \includegraphics[width=\linewidth]{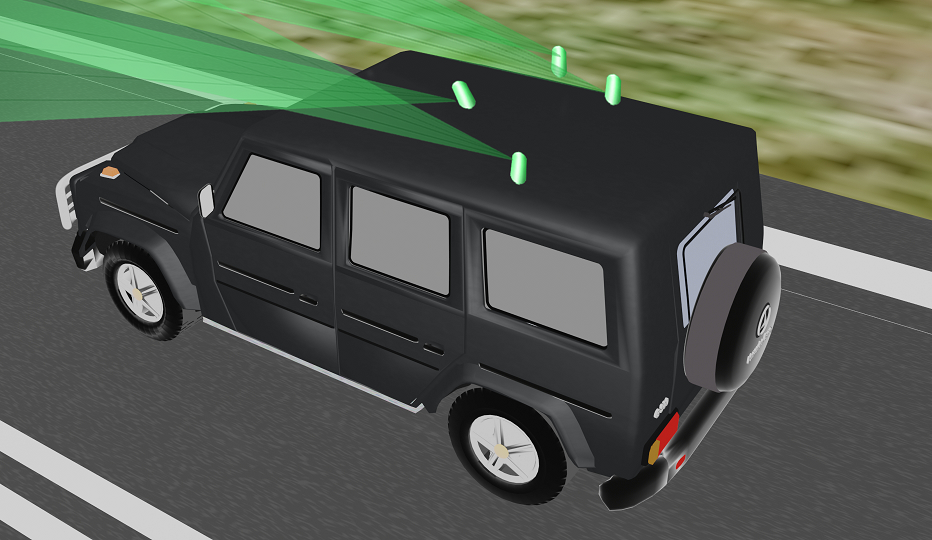}
                \caption{Pyramid-pitch}
    \end{subfigure}%
\vspace*{-3mm}
  \caption{Illustrations of different multi-LiDAR placements used in the experiments}
  \label{fig:lidar-placement}
   \vspace*{-5mm}
\end{figure*}

The ego-vehicle has its coordinate frame at its geometric center at $[40,20,0,0,0]$ with respect to the ROI frame of reference. All LiDAR configurations are illustrated in Figure \ref{fig:lidar-placement}, and their detailed coordinates are given in Table {\ref{detailed-coords}}. These coordinates are with respect to the ego vehicle's coordinate frame and will be transformed to ROI framework, as shown in Figure {\ref{fig:all-frames}}.  Note that the coordinate frames in CARLA are left-handed, and we make the extra transformation in ROI when calculating the POG and surrogate metric.

\subsection{Training Details of Detection Algorithms}
We use representative algorithms from OpenPCDet \cite{openpcdet2020} to evaluate the performance under different LiDAR placements. We keep all the models' hyperparameters the same as the default KITTI configuration files and change the optimization parameters to fine-tune the pre-trained models using the collected data from different LiDAR placements in CARLA.
Details of the optimization hyper-parameters are given in Table {\ref{table:hyper-param}}. Since only the front half of the point cloud is used to fine-tune the detection models, we change \texttt{POINT\_CLOUD\_RANGE} to be $[0, -20, -3, 40, 20, 1]$ as well. We ensure that the hyper-parameters are the same for all experiments to fairly compare the detection performance, and the detection performance is tested at epoch 10 for all models.

\begin{table}[ht!]
\centering
\begin{tabular}{|c||c|c|c|c|c|}
\hline Placement
 & x & y & z & roll & pitch \\ \hline
 \multirow{4}{*}{Line} &
\multirow{4}{*}{\makecell{-0.0 \\ 0.0 \\   0.0 \\ 0.0}}  &  \multirow{4}{*}{\makecell{  -0.6 \\ -0.4 \\ 0.4 \\ 0.6}}  &  \multirow{4}{*}{\makecell{  2.2 \\ 2.2  \\   2.2 \\ 2.2}}  &  \multirow{4}{*}{\makecell{  0.0 \\ 0.0  \\   0.0 \\ 0.0}}  &  \multirow{4}{*}{\makecell{  0.0 \\ 0.0  \\   0.0 \\ 0.0}}  \\
& & & & & \\
& & & & & \\
& & & & & \\ \hline
\multirow{4}{*}{Center} &
\multirow{4}{*}{\makecell{0.0  \\ 0.0  \\ 0.0 \\ 0.0}}  &  \multirow{4}{*}{\makecell{0.0  \\ 0.0  \\ 0.0 \\ 0.0}}  &  \multirow{4}{*}{\makecell{2.4  \\ 2.6  \\ 2.8 \\ 3.0}}  &  \multirow{4}{*}{\makecell{0.0  \\ 0.0 \\ 0.0 \\ 0.0}}  &  \multirow{4}{*}{\makecell{0.0  \\ 0.0  \\ 0.0 \\ 0.0}}  \\
& & & & & \\
& & & & & \\
& & & & & \\ \hline
\multirow{4}{*}{Trapezoid} &
\multirow{4}{*}{\makecell{-0.4\\ -0.4\\ 0.2\\ 0.2}} &  \multirow{4}{*}{\makecell{  0.2\\ -0.2\\ 0.5\\ -0.5}} &  \multirow{4}{*}{\makecell{  2.2  \\   2.2 \\ 2.2 \\   2.2}} &  \multirow{4}{*}{\makecell{  0.0  \\   0.0 \\ 0.0 \\   0.0}} &  \multirow{4}{*}{\makecell{  0.0  \\   0.0 \\ 0.0 \\   0.0}} \\
& & & & & \\
& & & & & \\
& & & & & \\ \hline
\multirow{4}{*}{Square} &
\multirow{4}{*}{\makecell{-0.5  \\ -0.5 \\ 0.5 \\   0.5}} &  \multirow{4}{*}{\makecell{  0.5  \\ - 0.5 \\ 0.5 \\ -0.5}} &  \multirow{4}{*}{\makecell{  2.2  \\   2.2 \\ 2.2 \\   2.2}} &  \multirow{4}{*}{\makecell{  0.0  \\   0.0 \\ 0.0 \\   0.0}} &  \multirow{4}{*}{\makecell{  0.0  \\   0.0 \\ 0.0 \\   0.0}} \\
& & & & & \\
& & & & & \\
& & & & & \\ \hline

 \multirow{4}{*}{Line-roll} &
\multirow{4}{*}{\makecell{- 0.0 \\ 0.0 \\   0.0 \\ 0.0}}  &  \multirow{4}{*}{\makecell{  -0.6 \\ -0.4 \\ 0.4 \\ 0.6}}  &  \multirow{4}{*}{\makecell{  2.2 \\ 2.2  \\   2.2 \\ 2.2}}  &  \multirow{4}{*}{\makecell{  -0.28 \\ 0.0  \\   0.0 \\ 0.28}}  &  \multirow{4}{*}{\makecell{  0.0 \\ 0.0  \\   0.0 \\ 0.0}}  \\
& & & & & \\
& & & & & \\
& & & & & \\ \hline
 \multirow{4}{*}{Pyramid} &
\multirow{4}{*}{\makecell{-0.2\\0.4\\ -0.2\\ -0.2}}  &  \multirow{4}{*}{\makecell{  -0.6\\ 0.0\\ 0.0\\ 0.6}}  &  \multirow{4}{*}{\makecell{  2.2\\ 2.4\\ 2.6\\ 2.2}}  &  \multirow{4}{*}{\makecell{  0.0 \\ 0.0  \\   0.0 \\ 0.0}}  &  \multirow{4}{*}{\makecell{  0.0 \\ 0.0  \\   0.0 \\ 0.0}}  \\
& & & & & \\
& & & & & \\
& & & & & \\ \hline
 \multirow{4}{*}{Pyramid-roll} &
\multirow{4}{*}{\makecell{-0.2\\0.4\\ -0.2\\ -0.2}}  &  \multirow{4}{*}{\makecell{  -0.6\\ 0.0\\ 0.0\\ 0.6}}  &  \multirow{4}{*}{\makecell{  2.2\\ 2.4\\ 2.6\\ 2.2}}  &  \multirow{4}{*}{\makecell{  -0.28 \\ 0.0  \\   0.0 \\ 0.28}}  &  \multirow{4}{*}{\makecell{  0.0 \\ 0.0  \\   0.0 \\ 0.0}}  \\
& & & & & \\
& & & & & \\
& & & & & \\ \hline
 \multirow{4}{*}{Pyramid-pitch} &
\multirow{4}{*}{\makecell{-0.2\\0.4\\ -0.2\\ -0.2}}  &  \multirow{4}{*}{\makecell{  -0.6\\ 0.0\\ 0.0\\ 0.6}}  &  \multirow{4}{*}{\makecell{  2.2\\ 2.4\\ 2.6\\ 2.2}}  &  \multirow{4}{*}{\makecell{  0.0 \\ 0.0  \\   0.0 \\ 0.0}}  &  \multirow{4}{*}{\makecell{  0.0 \\ -0.09  \\   0.0 \\ 0.0}}  \\
& & & & & \\
& & & & & \\
& & & & & \\ \hline
\end{tabular}
\caption{Coordinates of LiDAR sensors with respect to the ego-vehicle coordinate frame. All values of \textit{x,y,z} are in meters and roll and pitch angles are in \textit{rad}.}
\label{detailed-coords}
\vspace*{-3mm}
\end{table}

\begin{table}[h]
\centering
\begin{tabular}{|c|c|}
\hline
Hyperparameter & Value \\ \hline
Epochs & 10 \\ \hline
Optimizer & adam\_onecycle \\ \hline
Learning Rate & 0.01 \\ \hline
Weight Decay: & 0.01 \\ \hline
Momentum: & 0.9 \\ \hline
Learning Rate Clip & 0.0000001 \\ \hline
Learning Rate Decay & 0.1 \\ \hline
Div Factor & 10 \\ \hline
Warmup Epoch & 1 \\ \hline
Learning Rate Warmup & False \\ \hline
Gradient Norm Clip & 10 \\ \hline
MOMS & {[}0.95, 0.85{]} \\ \hline
PCT\_START & 0.1 \\ \hline
\end{tabular}
\caption{Hypeparameters for optimization in model training}
\label{table:hyper-param}
\end{table}

\begin{table*}[]
\centering
\resizebox{0.9\textwidth}{!}{
\begin{tabular}{|c||c|c|c|c||c|c|c|c|}
\hline
\multirow{2}{*}{Models}          & \multicolumn{4}{c||}{ Recall rcnn @0.50 IoU}                                          & \multicolumn{4}{c|}{Recall rcnn @0.70 IoU}                                 \\ \cline{2-9}
                                    & Center & Line & Pyramid & Trapezoid   & Center & Line &  Pyramid & Trapezoid \\ \hline
PV-RCNN  \cite{shi2020pv}
     &     0.5834&	0.6009&	0.6260&	0.6103
 &	0.4192&	0.4330&	0.4610&	0.4404
                    \\ \hline
Voxel RCNN  \cite{deng2020voxel}
&0.5686&	0.5858&	0.6112&	0.5929
   & 0.3901&	0.4093&	0.4299&	0.4172
                                                 \\ \hline
PointRCNN \cite{Shi_2019_CVPR}
&0.4462&	0.4584&	0.4722&	0.4593
& 0.3030&	0.3346&	0.3321&	0.3392
                 \\ \hline
PointRCNN-IoU \cite{Shi_2019_CVPR}
&	0.4437&	0.4633&	0.4706&	0.4597
&0.3020&	0.3300&	0.3325&	0.3373
                         \\ \hline
SECOND \cite{yan2018second}
& 0.4590&	0.4792&	0.5035&	0.4835
&0.2827&	0.2960&	0.3104&	0.3050
                                \\ \hline
SECOND-IoU  \cite{yan2018second}
&	0.5788&	0.5944&	0.6249&	0.6052
 &	0.3726&	0.3838&	0.4107&	0.3966
                                 \\ \hline
\end{tabular}}
\caption{Comparison of recall performance under various LiDAR configurations  using different algorithms.
}
\label{tab:recall_compare}
\vspace*{-3mm}
\end{table*}

\section{Appendix - More Experimental Results and Analysis}
\label{app:more_ressults}
\subsection{Recall under Different LiDAR Configurations}
Besides the average precision to show the detection performance, we consider using the overall recall (rcnn) with IoU of 0.5 and 0.7 to show the detection performance of all the objects and validate the relationship between our surrogate metric. The comparison of recall performance under different LiDAR placements can be found in Table \ref{tab:recall_compare}. We can see the recall metric varies a lot under different LiDAR placements for the same detection models. Specifically, \textit{Pyramid} almost gets the best performance for all the algorithms with IoU of both 0.5 and 0.7, which is different from the performance of average precision in Table \ref{tab:expt1}.

\begin{figure*}[t]
  \centering
  \begin{subfigure}[b]{0.48\textwidth}
  \includegraphics[width=\linewidth]{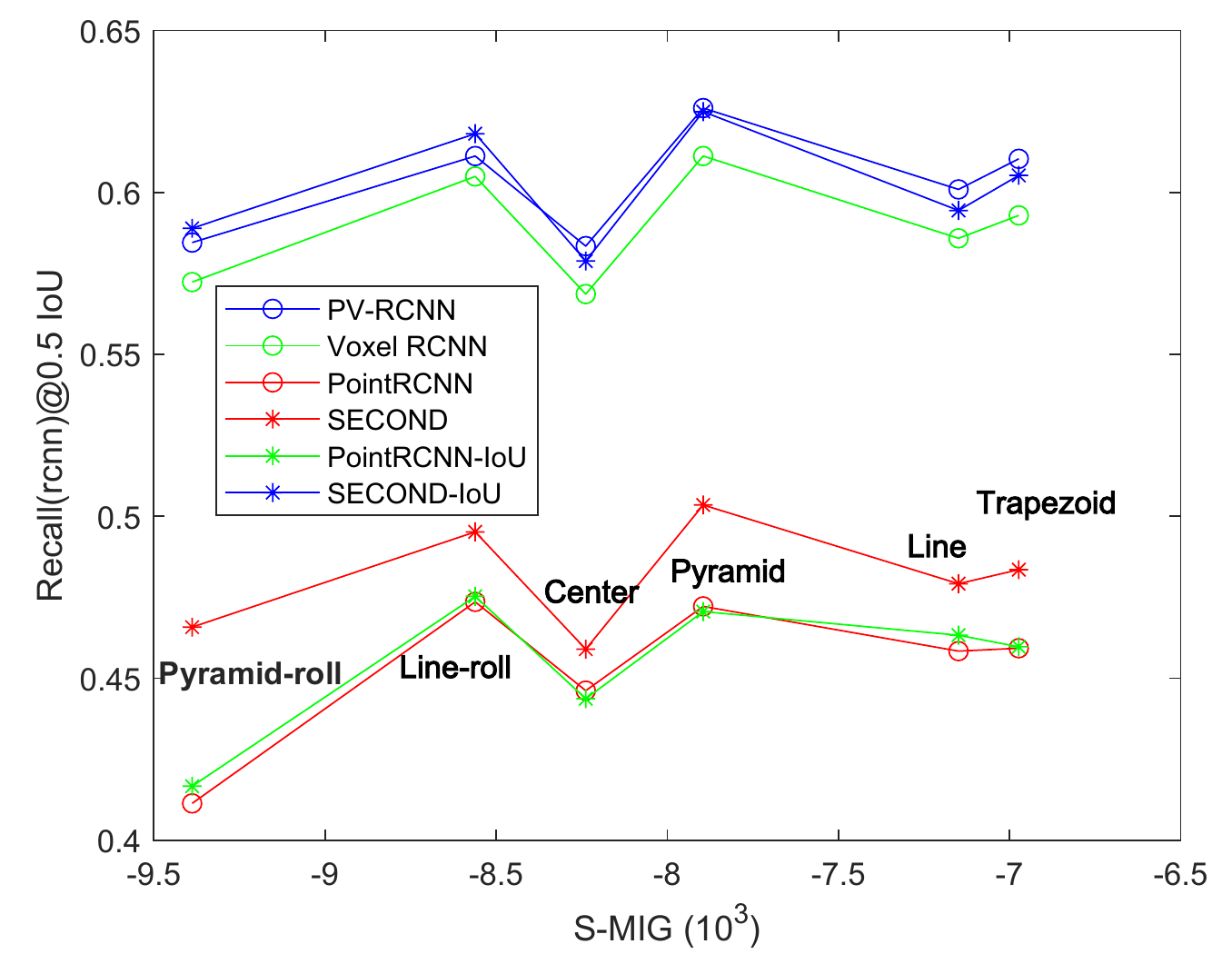}
    \end{subfigure}%
    \begin{subfigure}[b]{0.48\textwidth}
  \includegraphics[width=\linewidth]{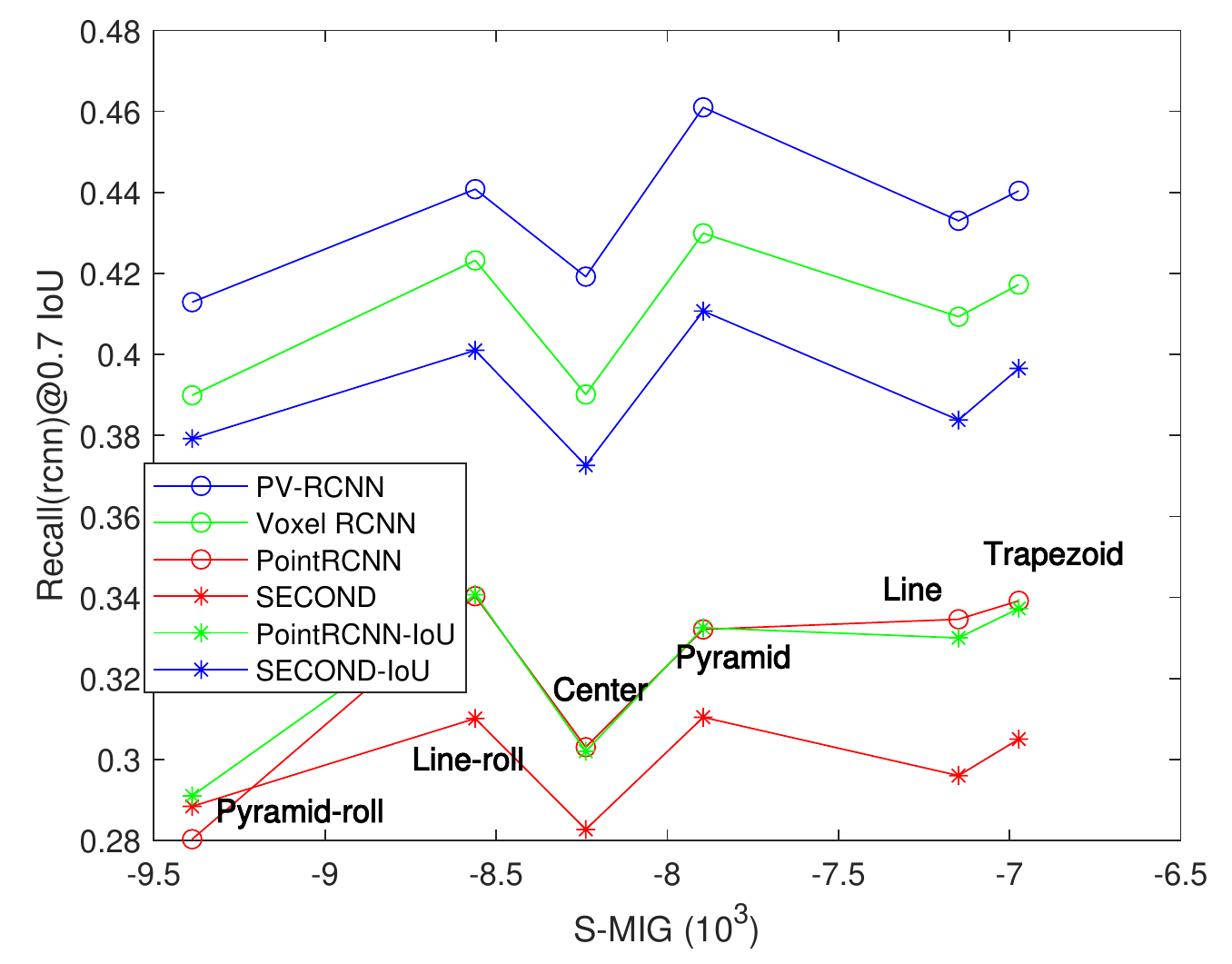}
    \end{subfigure}%
\vspace*{-3mm}
  \caption{The relationship overall recall performance  and overall surrogate metric of different LiDAR placements for  Car, Van and Cyclist.}
  \label{fig:recall_relation}
\end{figure*}

From Figure \ref{fig:recall_relation}, we can find the increasing trend between recall and our surrogate metric as well. Note that since the recall is calculated for all Cars, Vans and Cyclists, our total information gain surrogate metric is the sum of S-MIG of Car, Van and Cyclist. Furthermore, it can be seen that the performance variance under different LiDAR placements does not decrease as the IoU is going less, showing that the influence of LiDAR placement is consistent with the recall metrics, as the surrogate metric shows.

\begin{table}[]
\centering
\resizebox{0.45\textwidth}{!}{
\begin{tabular}{|c|c|c|c|}
\hline
3D  AP  & PV-RCNN\cite{shi2020pv}  &  PointRCNN\cite{Shi_2019_CVPR}  & \makecell[c]{S-MIG\\ $(10^3)$}  \\ \hline
Pyramid &    57.44
    &   44.28
       &   -5.64       \\ \hline
Pyramid-pitch &     53.46
     &     37.27
      &    -5.71      \\ \hline
\hline
BEV AP   & PV-RCNN\cite{shi2020pv}   & PointRCNN\cite{Shi_2019_CVPR}  & \makecell[c]{S-MIG\\ $(10^3)$} \\ \hline
Pyramid &  65.81
       &    56.97
      &    -5.64     \\ \hline
Pyramid-pitch &     62.33
    &     51.39
      &   -5.71     \\ \hline
\end{tabular}}
\caption{Influence of pitch rotation of front LiDARs on Car AP detection performance.
}
\label{tab:pitch_AP}
\vspace*{-5mm}
\end{table}

\subsection{Sensitiveness Analysis of Detection Algorithms}
In this section, we analyze how sensitive current different LiDAR-based detraction algorithms are to the influence of LiDAR placements. From  Figure \ref{fig:expt2} and Figure \ref{fig:recall_relation}, it can be found that point-based methods, like \textit{PointRCNN} and \textit{PointRCNN-IoU}, are sensitive to different LiDAR placements and have a relatively clear linear relationship with our surrogate metric. On the contrary, the detection performance of voxel-based methods fluctuates with different LiDAR configurations as well, but the linear relationship is less obvious, which is because point-based methods rely on the original point data collected from LiDAR and are highly related to the point cloud distribution and uncertainty revealed by our surrogate metric.

Besides, there are some detection algorithms where the fluctuation caused by LiDAR placement is even more significant than the difference between different algorithms given any LiDAR placement. Specifically, the recall with 0.7 IoU of \textit{Second} is better than \textit{PointRCNN} under \textit{Pyramid-roll}, while  \textit{Second} performs worse than \textit{PointRCNN}  using data collected under other LiDAR placements, showing that LiDAR configuration is also a critical factor in object detection. Therefore, there is still room to improve the 3D detection performance from LiDAR placement.

\begin{figure*}[t]
  \centering
    \begin{subfigure}[b]{0.23\textwidth}
  \includegraphics[width=\linewidth]{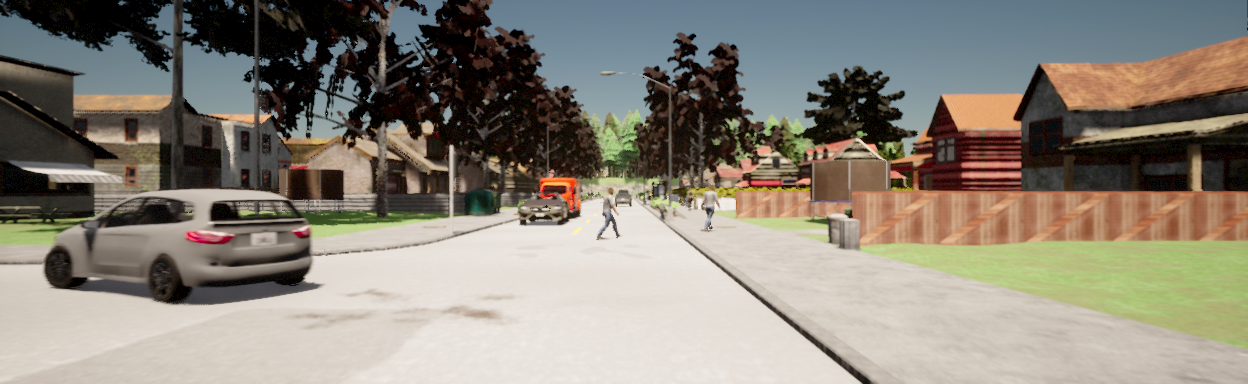}
    \end{subfigure}%
    \begin{subfigure}[b]{0.23\textwidth}
  \includegraphics[width=\linewidth]{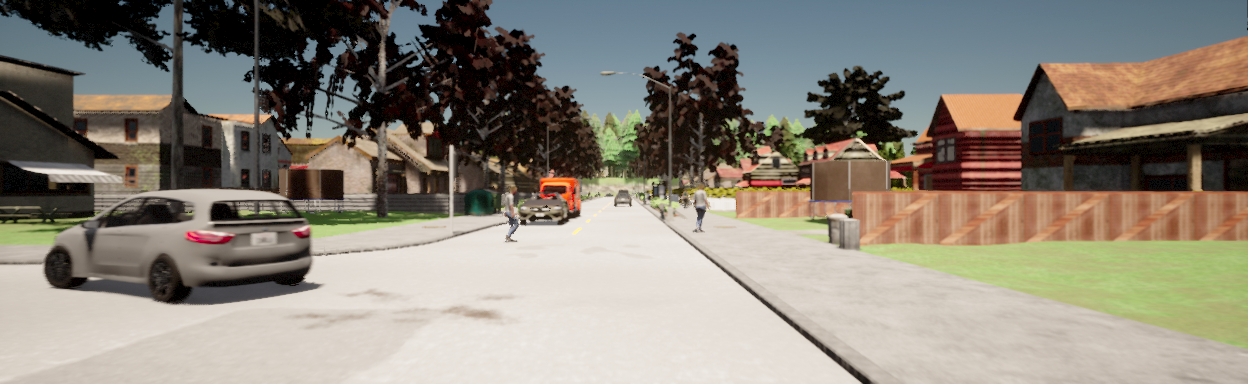}
    \end{subfigure}%
    \begin{subfigure}[b]{0.23\textwidth}
  \includegraphics[width=\linewidth]{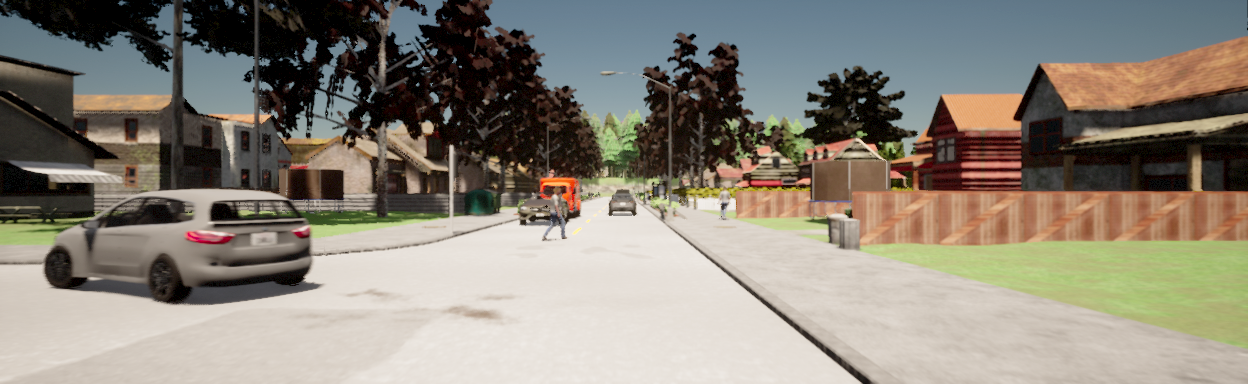}
    \end{subfigure}%
    \begin{subfigure}[b]{0.23\textwidth}
  \includegraphics[width=\linewidth]{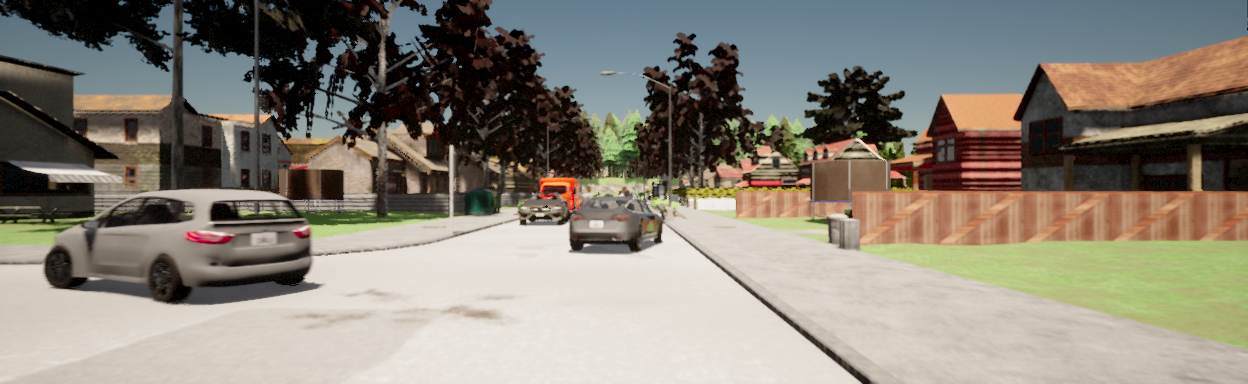}
    \end{subfigure}%

        \begin{subfigure}[b]{0.23\textwidth}
  \includegraphics[width=\linewidth,height=.58\linewidth]{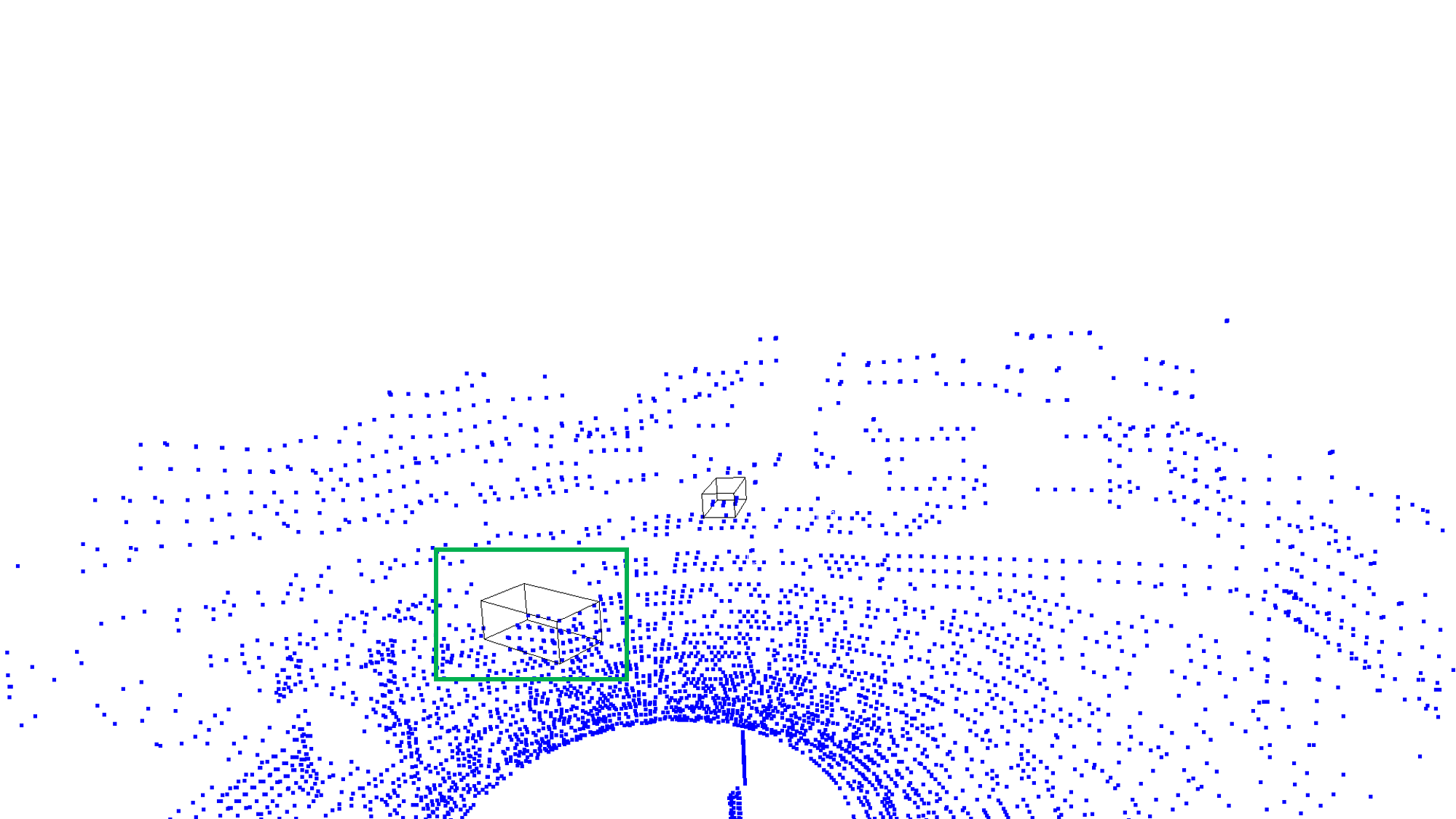}
    \end{subfigure}%
    \begin{subfigure}[b]{0.23\textwidth}
  \includegraphics[width=\linewidth]{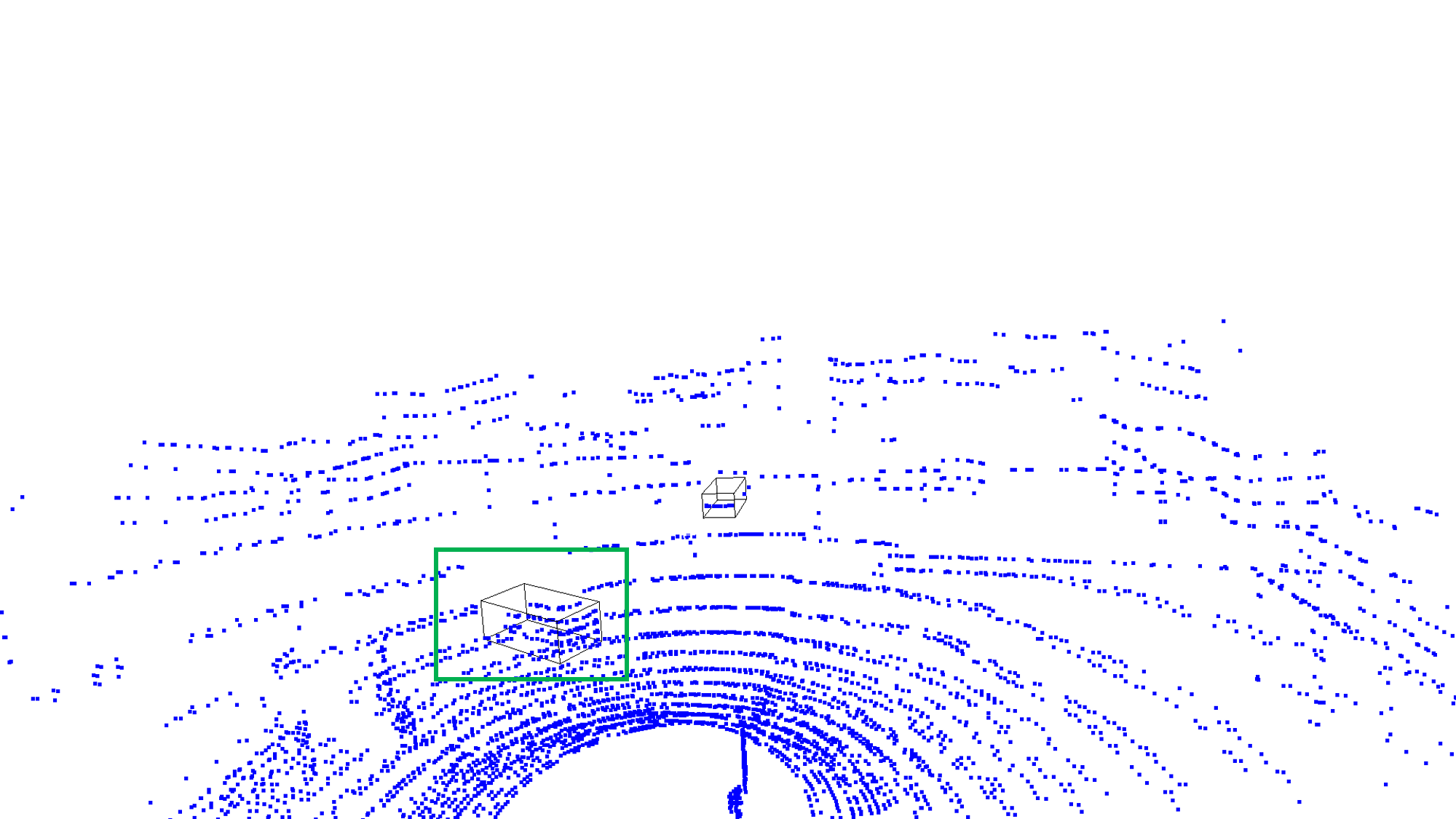}
    \end{subfigure}%
    \begin{subfigure}[b]{0.23\textwidth}
  \includegraphics[width=\linewidth]{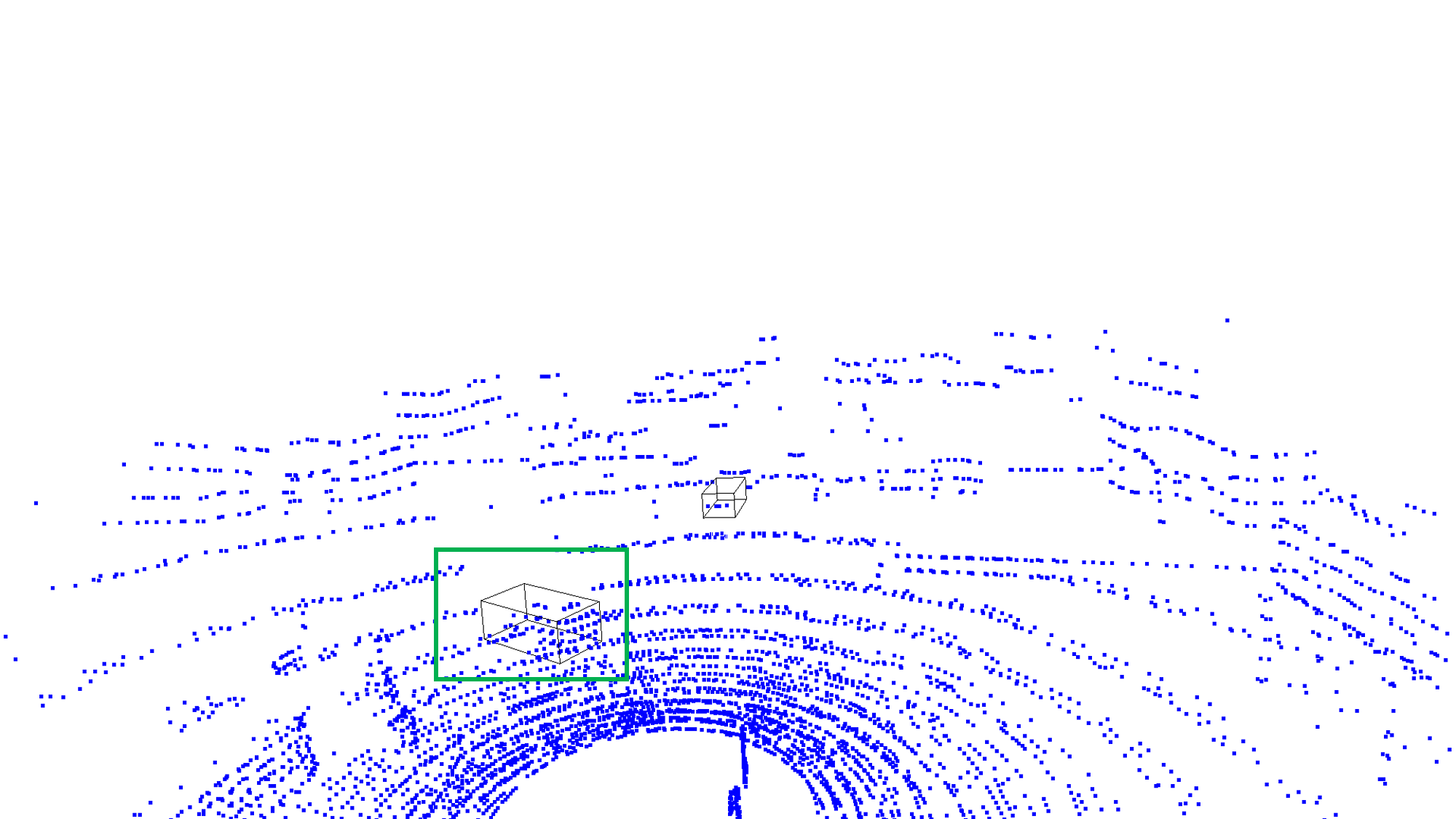}
    \end{subfigure}%
    \begin{subfigure}[b]{0.23\textwidth}
  \includegraphics[width=\linewidth]{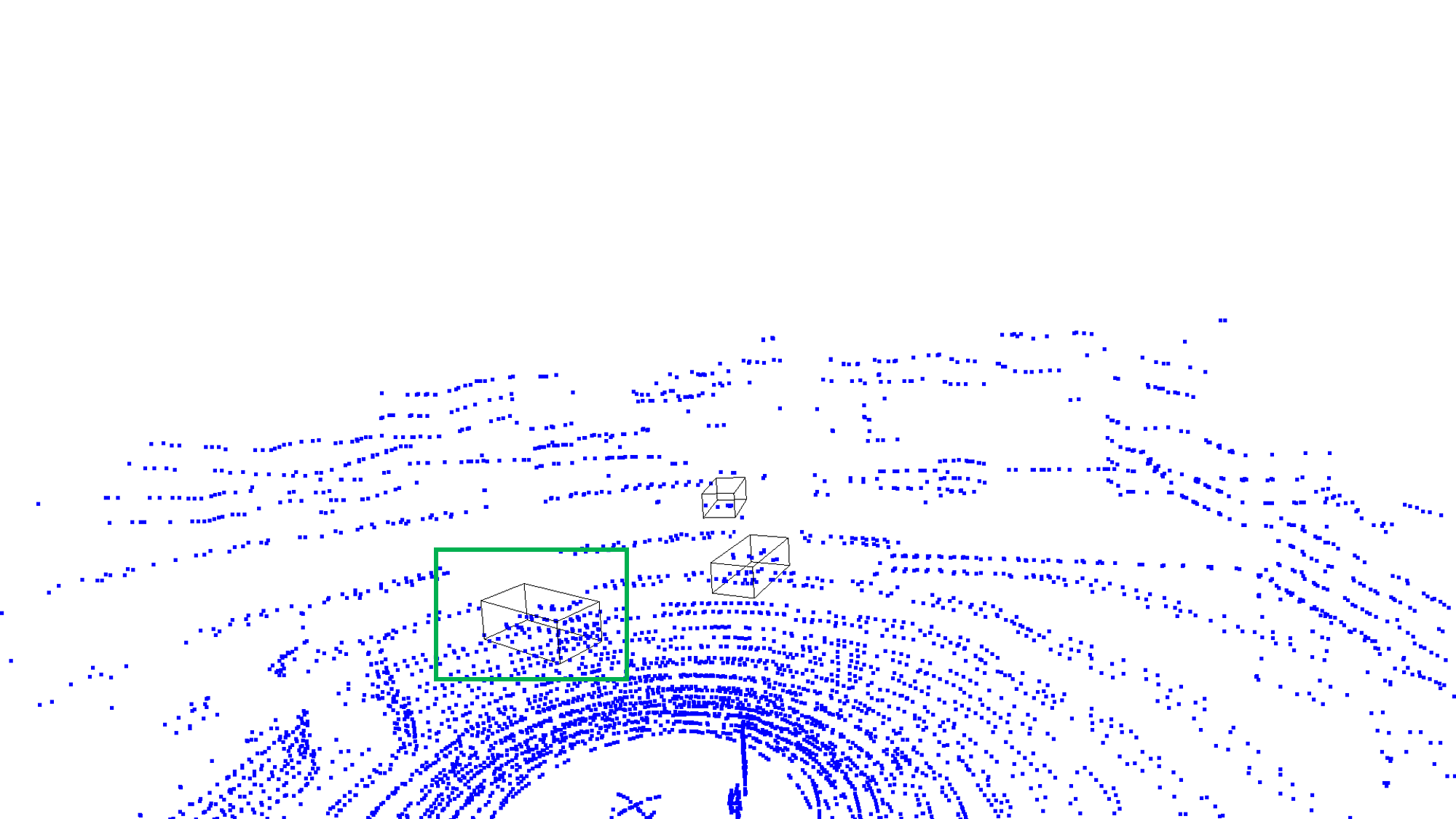}
    \end{subfigure}%

        \begin{subfigure}[b]{0.23\textwidth}
  \includegraphics[width=\linewidth]{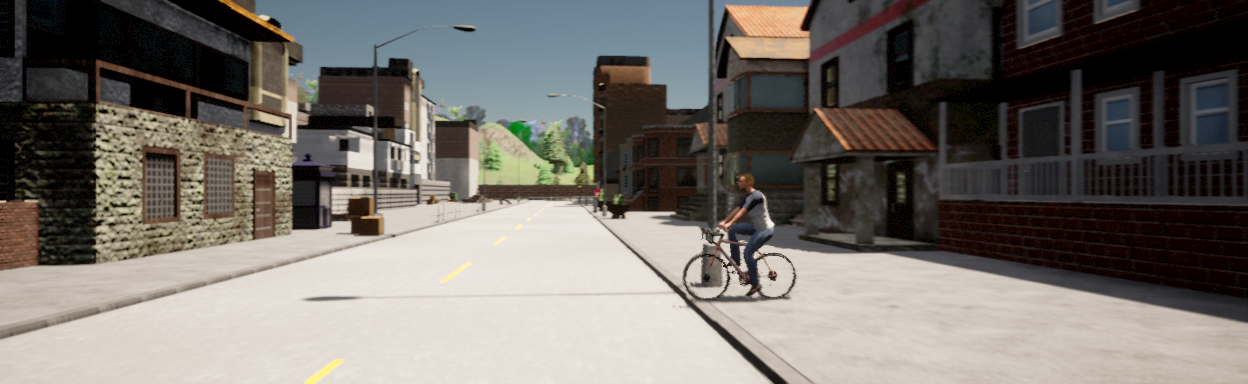}
    \end{subfigure}%
    \begin{subfigure}[b]{0.23\textwidth}
  \includegraphics[width=\linewidth]{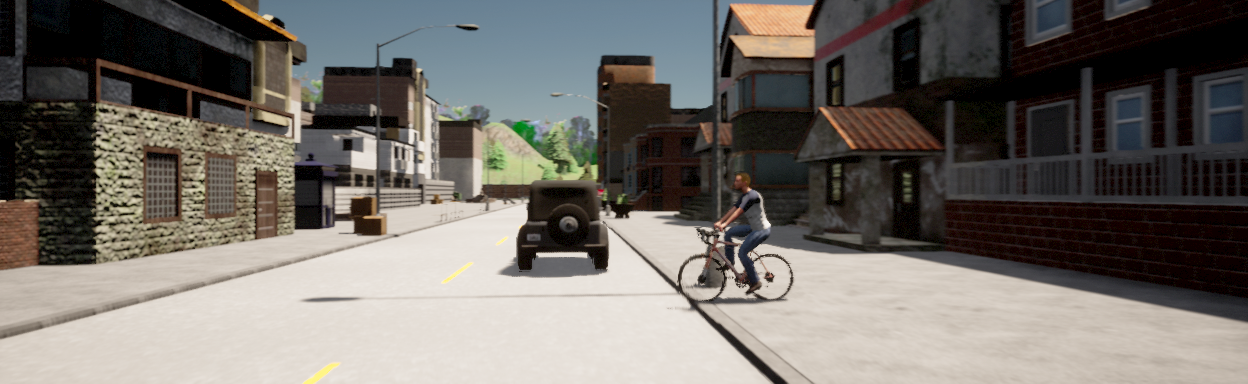}
    \end{subfigure}%
    \begin{subfigure}[b]{0.23\textwidth}
  \includegraphics[width=\linewidth]{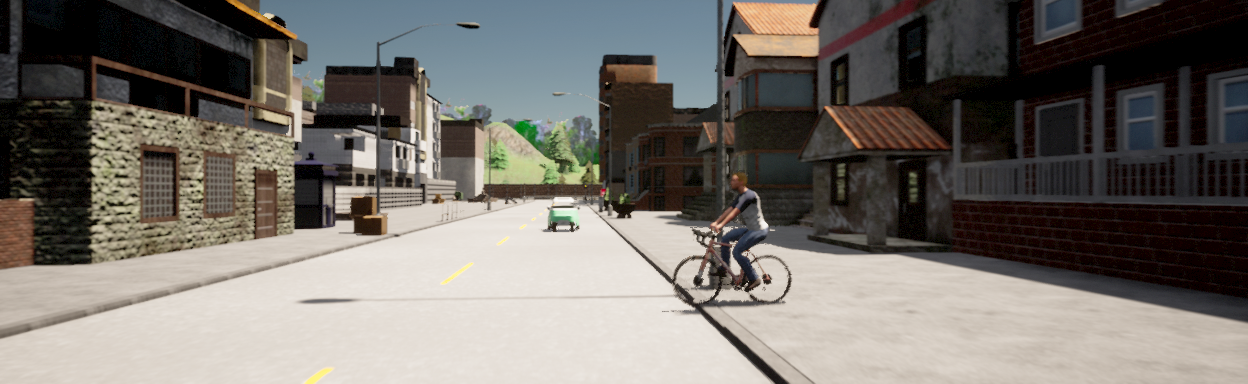}
    \end{subfigure}%
    \begin{subfigure}[b]{0.23\textwidth}
  \includegraphics[width=\linewidth]{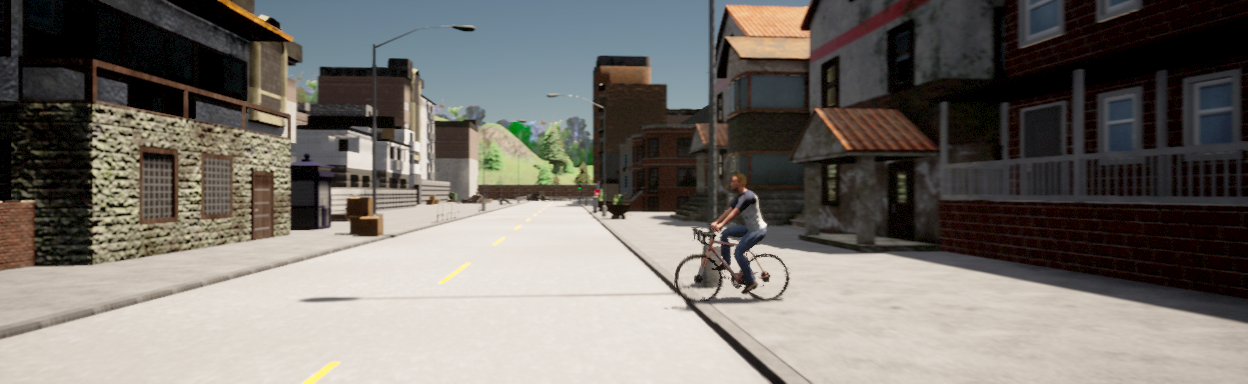}
    \end{subfigure}%

        \begin{subfigure}[b]{0.23\textwidth}
  \includegraphics[width=\linewidth,height=.58\linewidth]{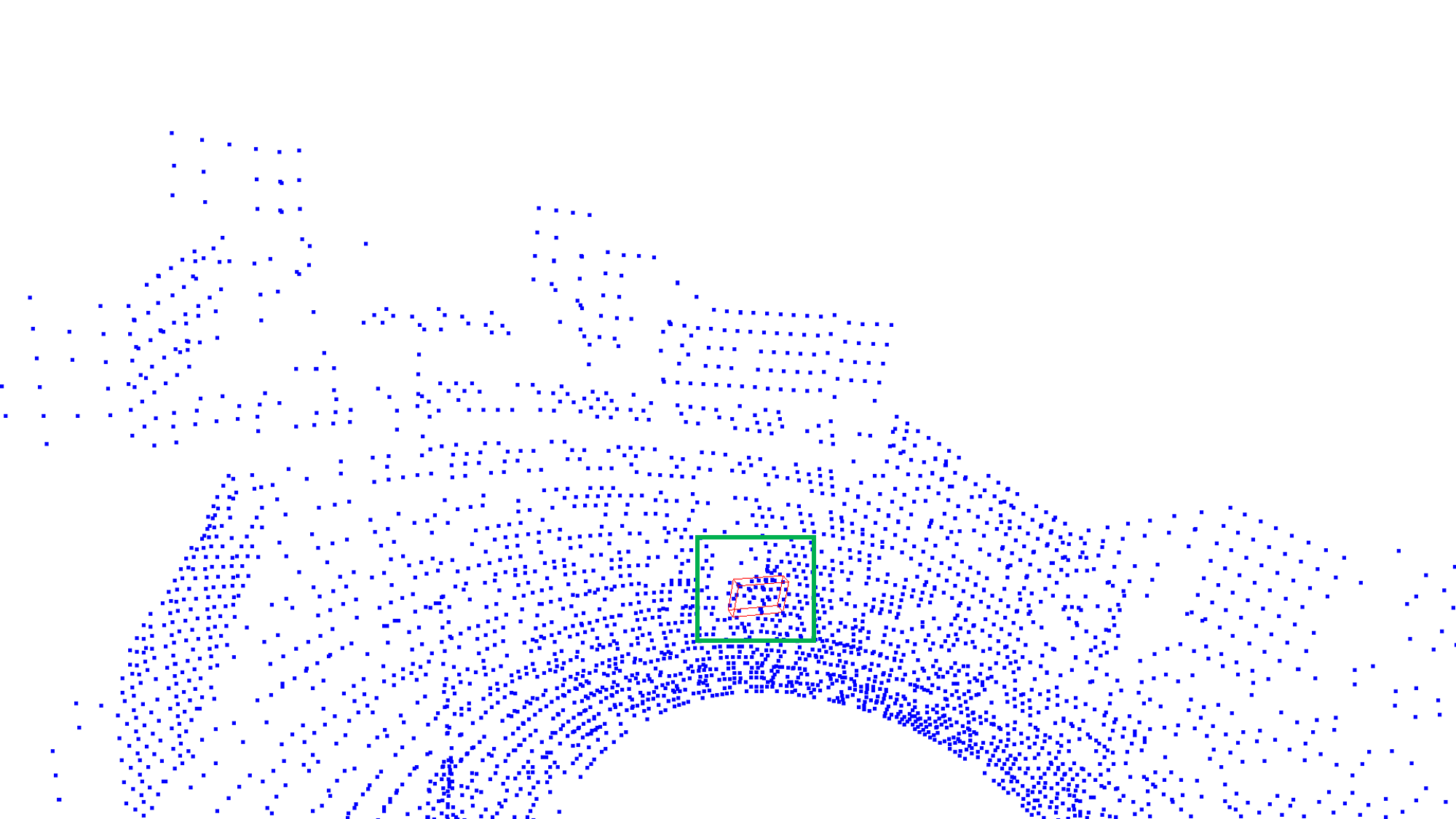}
    \end{subfigure}%
    \begin{subfigure}[b]{0.23\textwidth}
  \includegraphics[width=\linewidth]{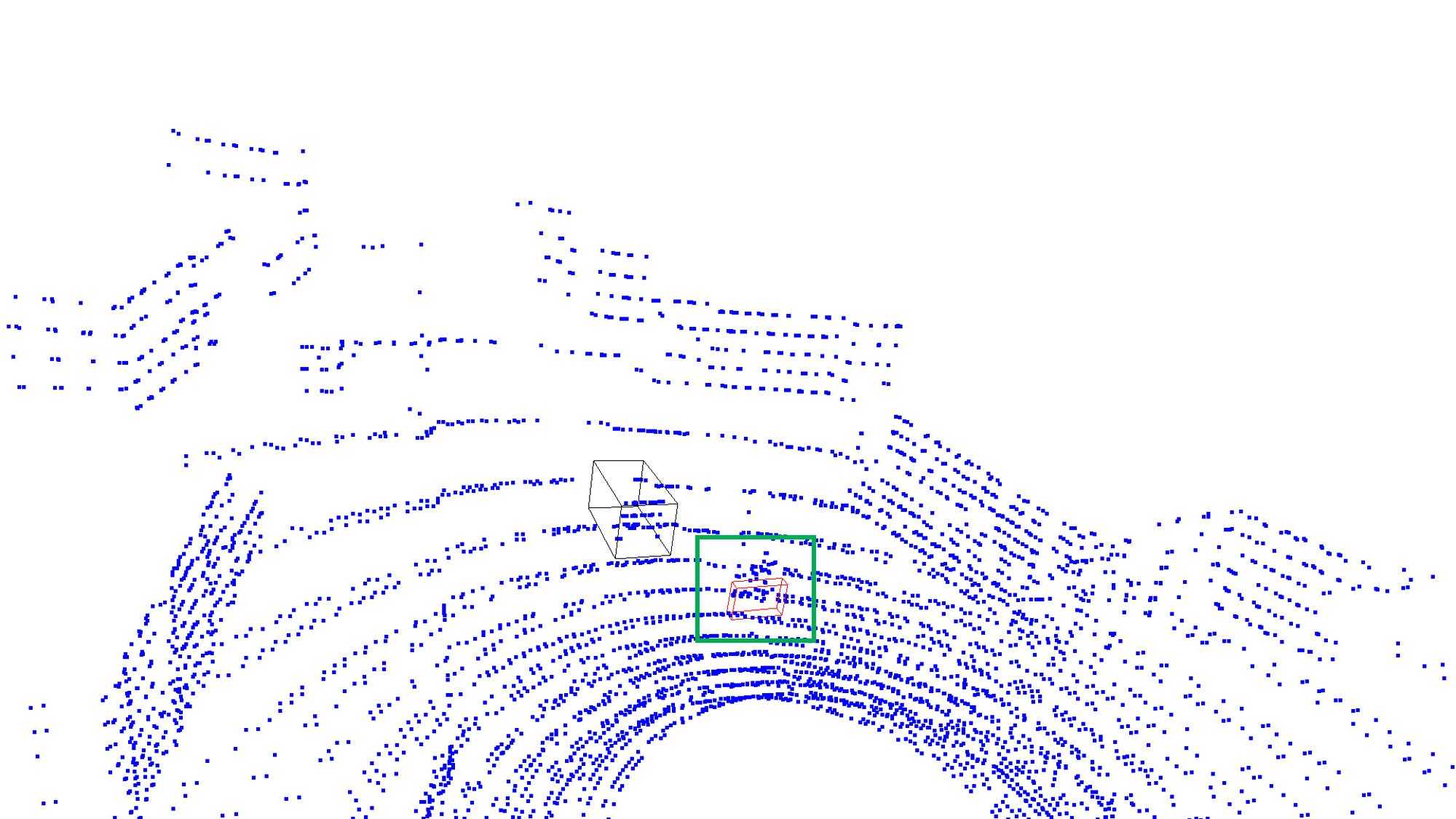}
    \end{subfigure}%
    \begin{subfigure}[b]{0.23\textwidth}
  \includegraphics[width=\linewidth]{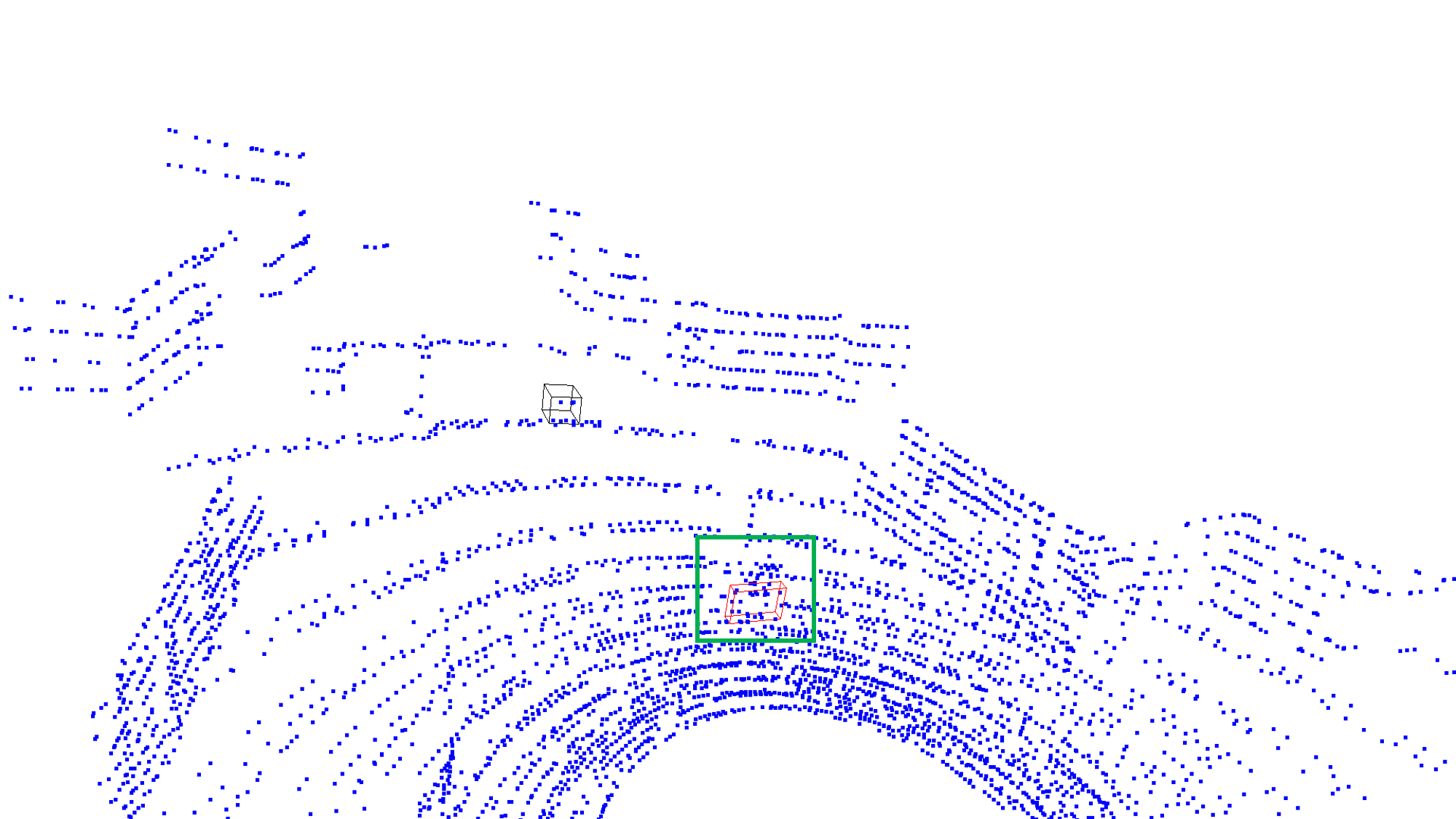}
    \end{subfigure}%
    \begin{subfigure}[b]{0.23\textwidth}
  \includegraphics[width=\linewidth]{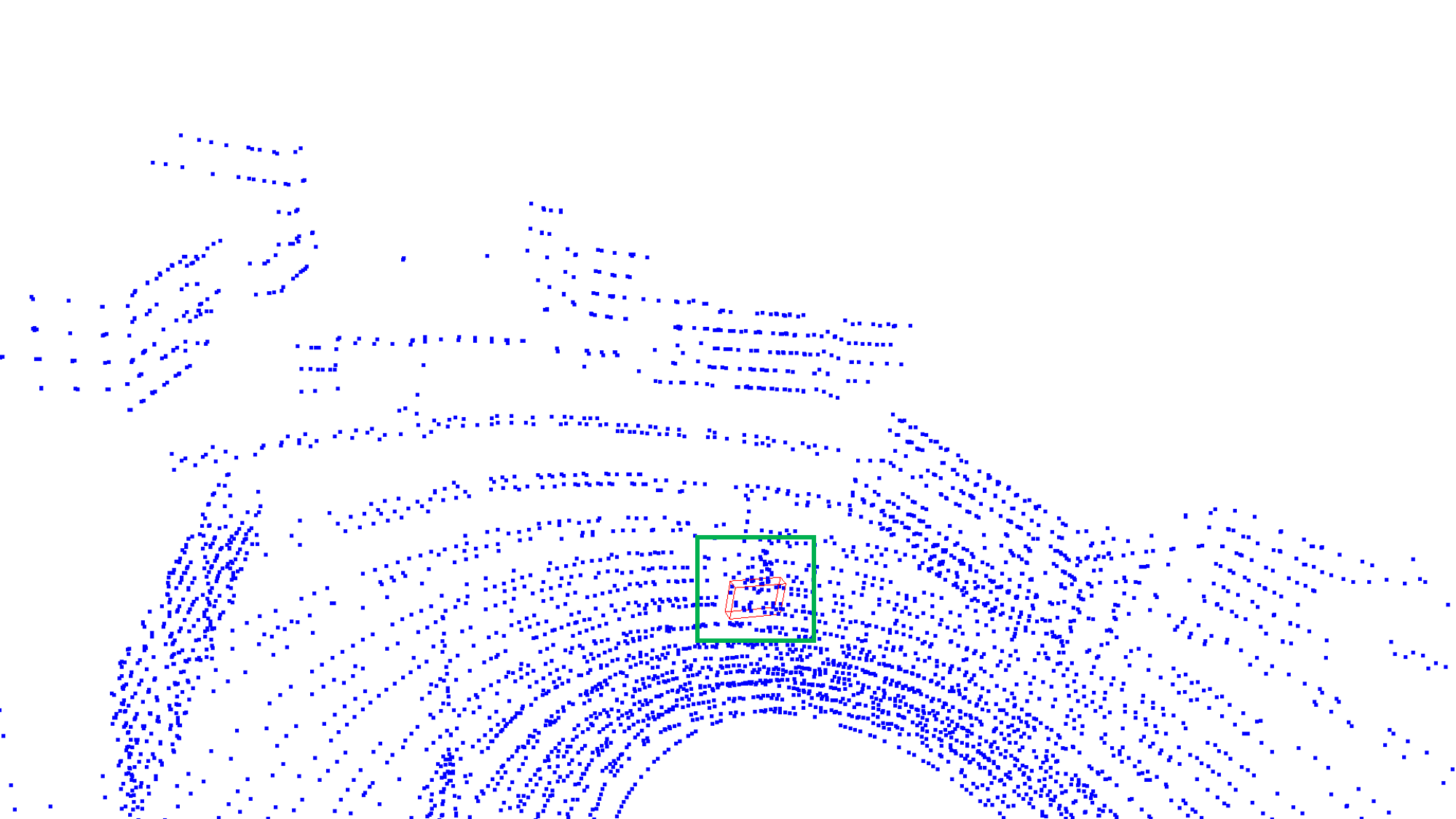}
    \end{subfigure}%

    \begin{subfigure}[b]{0.23\textwidth}
  \includegraphics[width=\linewidth,height=.58\linewidth]{images/baseline-center.png}
                \caption{Center}
    \end{subfigure}%
    \begin{subfigure}[b]{0.23\textwidth}
  \includegraphics[width=\linewidth,height=.58\linewidth]{images/line_norot.png}
                \caption{Line}
    \end{subfigure}%
    \begin{subfigure}[b]{0.23\textwidth}
  \includegraphics[width=\linewidth]{images/trapezoid.png}
                \caption{Trapezoid}
    \end{subfigure}%
    \begin{subfigure}[b]{0.23\textwidth}
  \includegraphics[width=\linewidth,height=.58\linewidth]{images/baseline-square.png}
                \caption{Square}
    \end{subfigure}%
  \caption{Visualization of point cloud distribution from different LiDAR configurations.}
  \label{fig:vis1}
   \vspace*{-3mm}
\end{figure*}

\begin{table}[]
\centering
\resizebox{0.45\textwidth}{!}{
\begin{tabular}{|c|c|c|c|}
\hline
Recall @0.5 IoU    & PV-RCNN\cite{shi2020pv}  &  PointRCNN\cite{Shi_2019_CVPR}  & \makecell[c]{S-MIG\\ $(10^3)$}  \\ \hline
Pyramid &    0.6260
    &   0.4722
       &   -7.90      \\ \hline
Pyramid-pitch &     0.6002
     &     0.4490
      &    -7.99      \\ \hline
\hline
Recall @0.7 IoU    & PV-RCNN\cite{shi2020pv}   & PointRCNN\cite{Shi_2019_CVPR}  & \makecell[c]{S-MIG\\ $(10^3)$} \\ \hline
Pyramid &  0.4610
       &    0.3321
      &    -7.90     \\ \hline
Pyramid-pitch &     0.4279
    &     0.3122
      &   -7.99     \\ \hline
\end{tabular}}
\caption{Influence of pitch rotation of front LiDARs on overall recall detection performance.
}
\label{tab:pitch_recall}
\vspace*{-5mm}
\end{table}

\subsection{Influence of Pitch Angles for Front LiDAR}
Along with the analysis of roll angle of sided LiDARs Section \ref{sec:ablation}, we also show the influence of pitch angle for front LiDAR, which is intuitively essential for object detection in the front 180-degree field of view. From Table \ref{tab:pitch_AP} and Table \ref{tab:pitch_recall}, it can be seen that the surrogate metric of \textit{Pyramid-pitch} placement is less than that of \textit{Pyramid} placement. The Car average precision and overall recall metrics are less, which shows that our surrogate can be used to evaluate the detection performance around the local neighborhood of LiDAR placement. Note that the surrogate metric is the sum of all the objects for recall comparison because the recall includes all Car, Van and Cyclist.

\begin{figure}[ht]
  \centering
    \begin{subfigure}[b]{0.16\textwidth}
  \includegraphics[width=\linewidth]{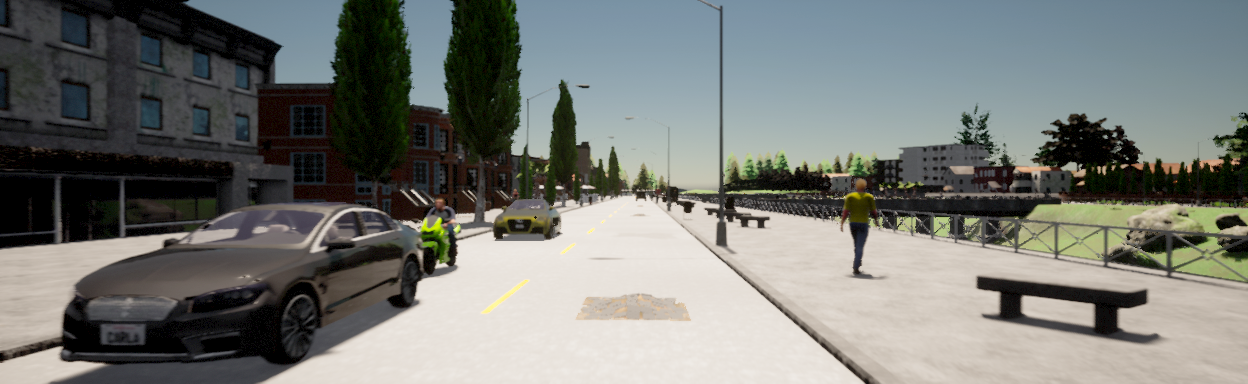}
    \end{subfigure}%
    \begin{subfigure}[b]{0.16\textwidth}
  \includegraphics[width=\linewidth]{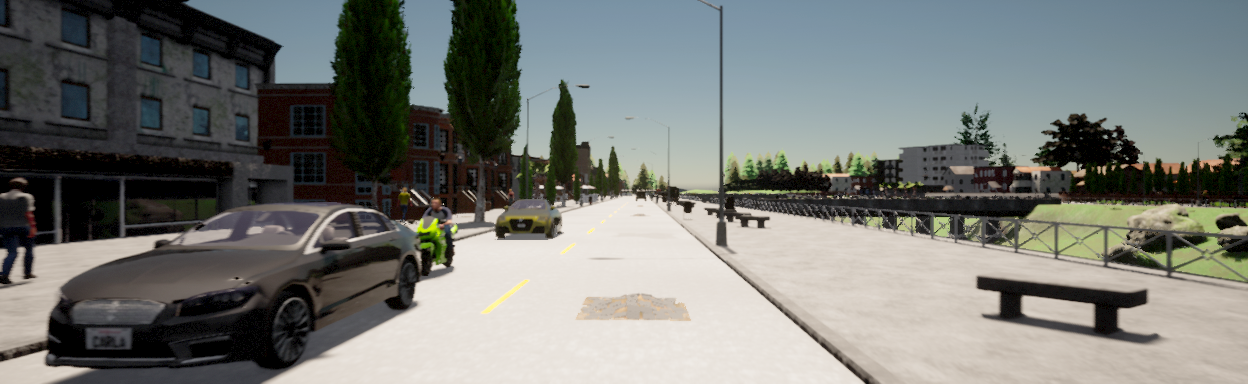}
    \end{subfigure}%
    \begin{subfigure}[b]{0.16\textwidth}
  \includegraphics[width=\linewidth]{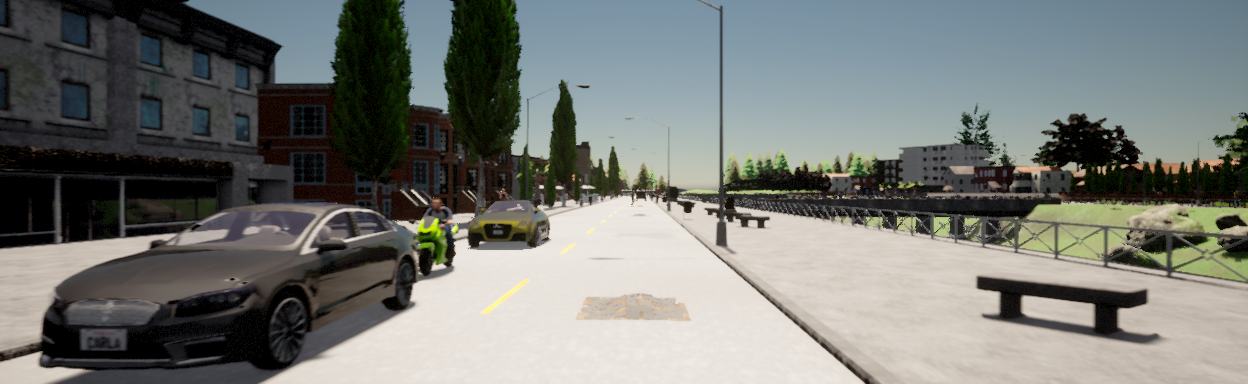}
    \end{subfigure}%

        \begin{subfigure}[b]{0.16\textwidth}
  \includegraphics[width=\linewidth,height=.58\linewidth]{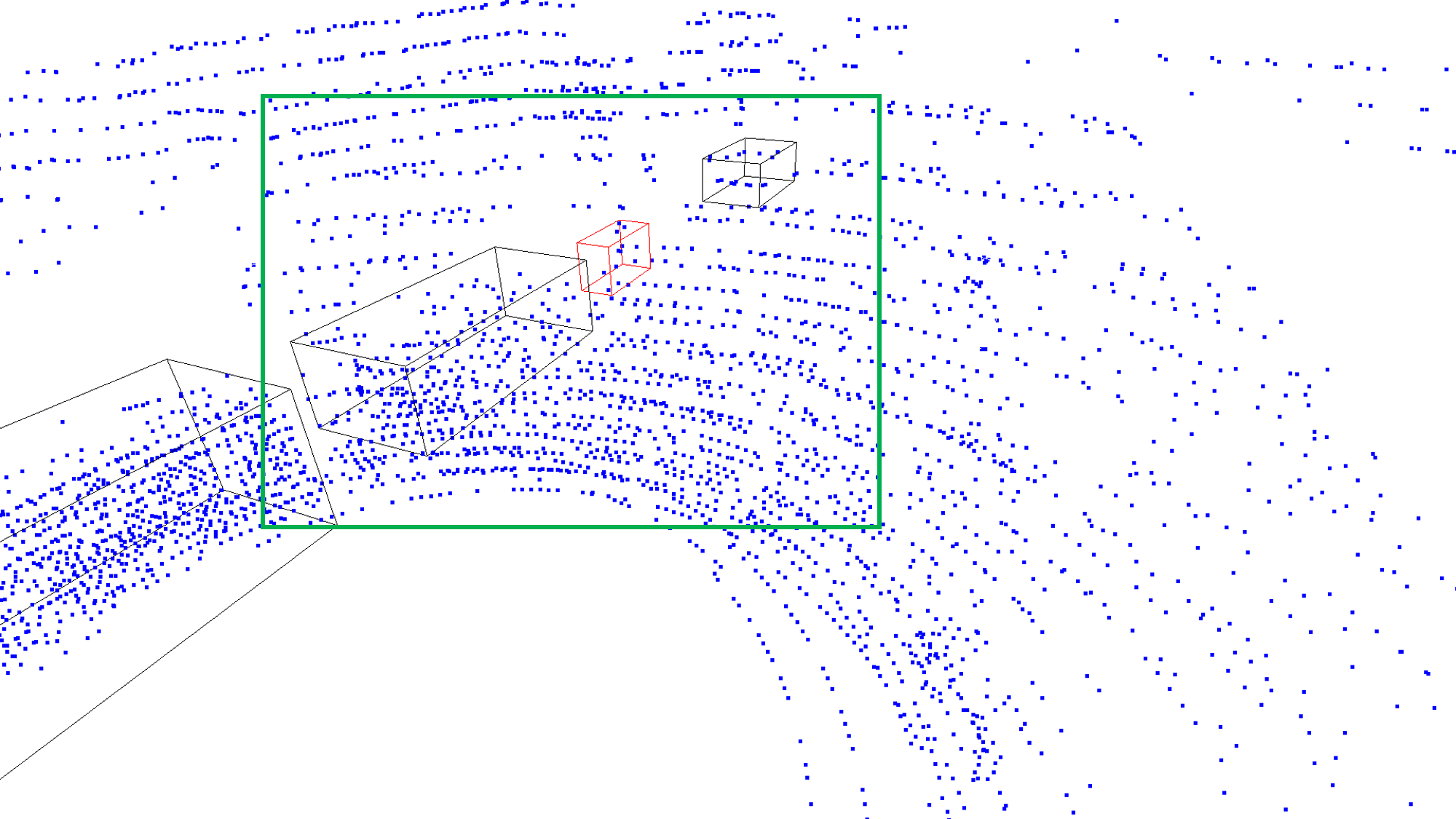}
    \end{subfigure}%
    \begin{subfigure}[b]{0.16\textwidth}
  \includegraphics[width=\linewidth]{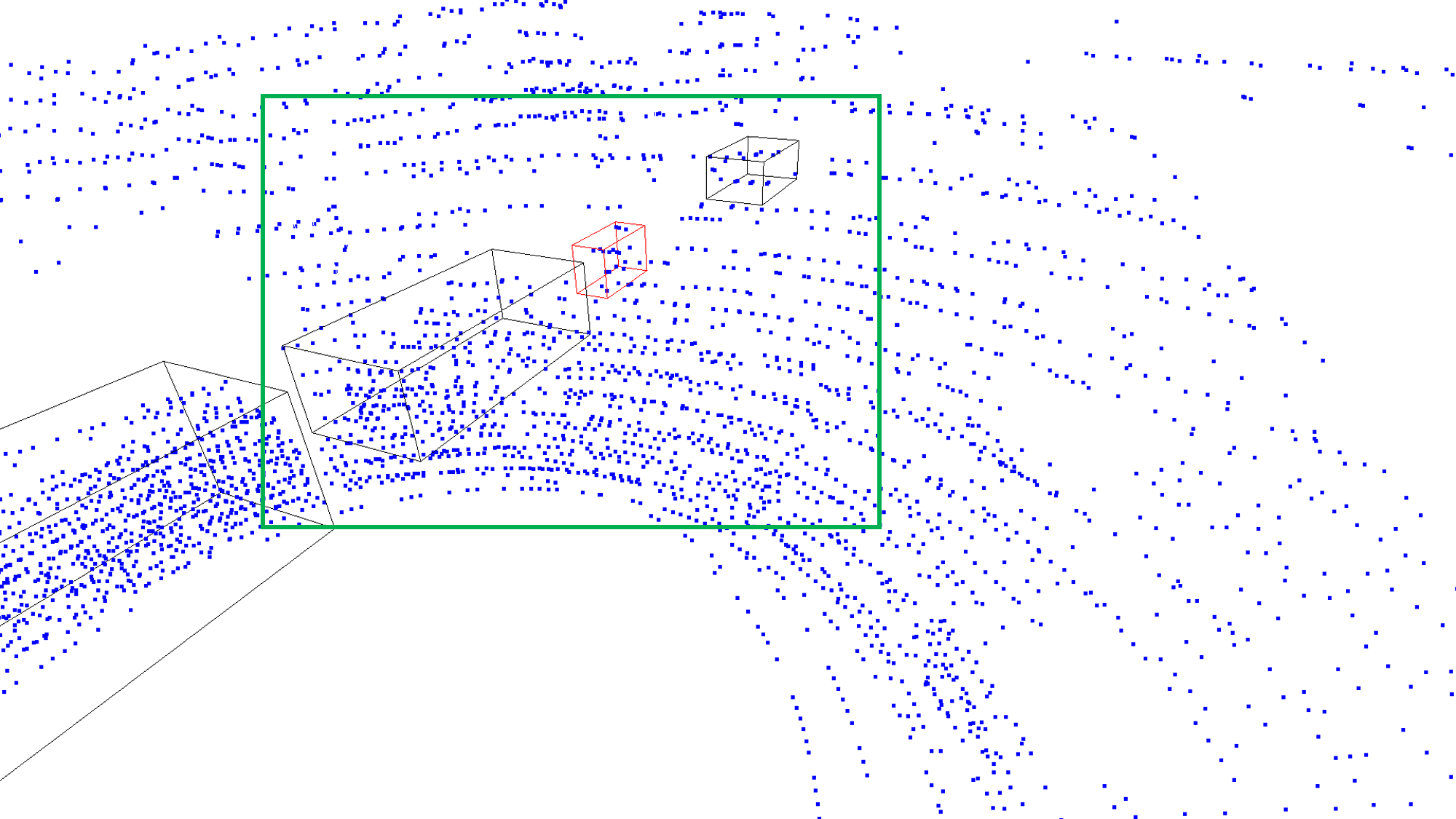}
    \end{subfigure}%
    \begin{subfigure}[b]{0.16\textwidth}
  \includegraphics[width=\linewidth]{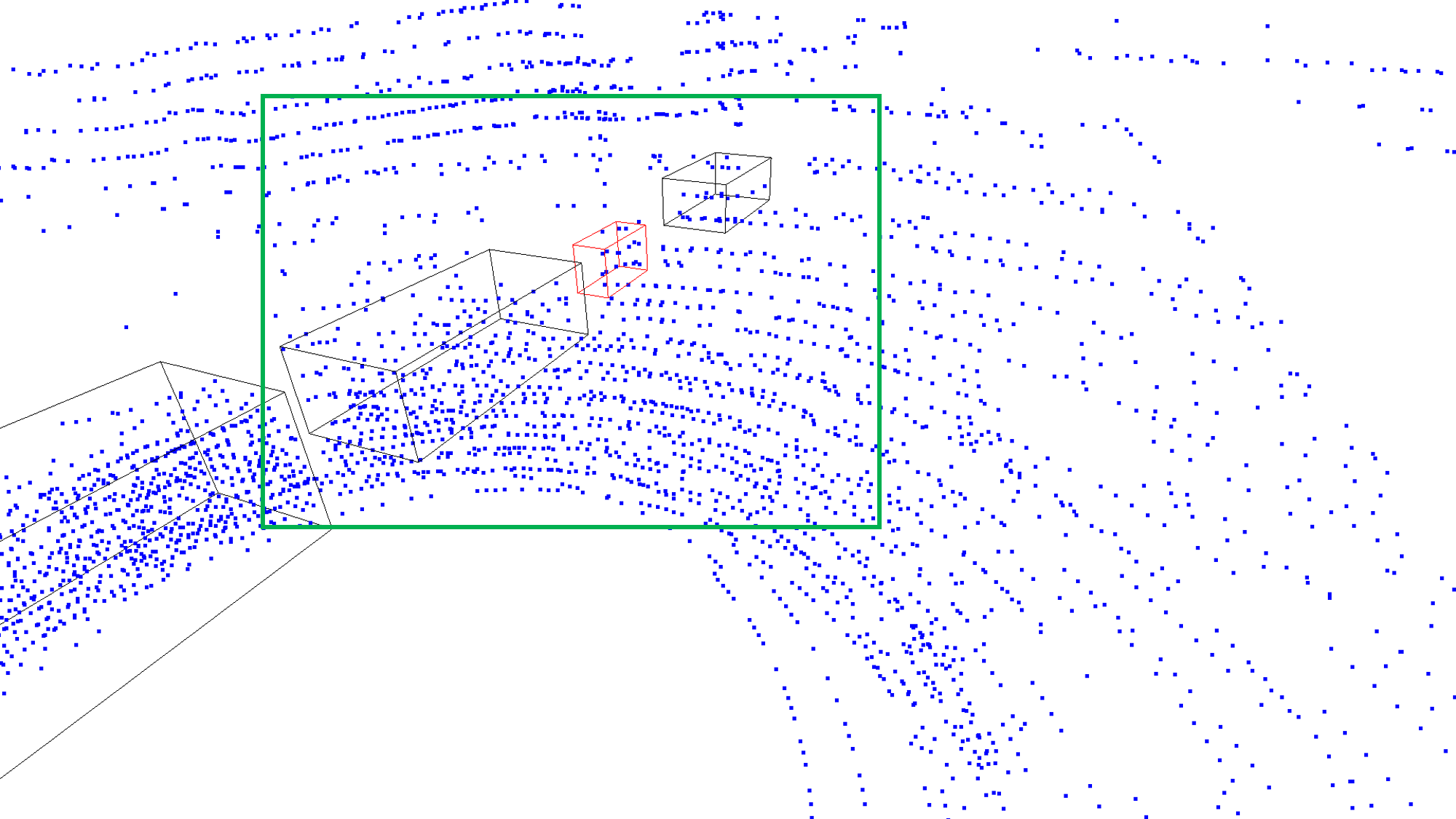}
    \end{subfigure}%

    \begin{subfigure}[b]{0.16\textwidth}
  \includegraphics[width=\linewidth,height=.58\linewidth]{images/triangle.png}
                \caption{Pyramid}
    \end{subfigure}%
    \begin{subfigure}[b]{0.16\textwidth}
  \includegraphics[width=\linewidth,height=.58\linewidth]{images/triangle_roll.png}
                \caption{Pyramid-roll}
    \end{subfigure}%
    \begin{subfigure}[b]{0.16\textwidth}
  \includegraphics[width=\linewidth]{images/triangle_pitch.png}
                \caption{Pyramid-pitch}
    \end{subfigure}%
  \caption{Comparison of point cloud distribution from LiDAR placement with roll and pitch angles.}
  \label{fig:vis2}
   \vspace*{-5mm}
\end{figure}

\subsection{LiDAR Placement for Pedestrain Detection}
Besides cars, vans and cyclists, e conducted an extra experiment considering the pedestrians mainly from the sidewalk as shown in the table below. The results further consolidate the significance of LiDAR placement to SOTA point cloud-based detection algorithms with the metric of AP at 0.5 IOU. It shows the \textit{Line} option almost performs the best due to its wide horizontal view field. Furthermore, we found that the influence on Pedestrian detection is even larger than Car and Cyclists as shown in Table \ref{tab:expt1} text of the paper, where the four SOTA models are affected by 30\% of \textit{Line} placement.

\begin{table}[H]
\centering
\resizebox{0.47\textwidth}{!}{
\begin{tabular}{|c||c|c|c|c|}
\hline
       3D  (AP@0.50)                             & Center & Line & Pyramid & Trapezoid  \\ \hline
PV-RCNN   \cite{shi2020pv}
      &   12.80 & \textbf{	16.35}& 	 10.95 & 	 13.57
                    \\ \hline
Voxel RCNN  \cite{deng2020voxel}
&    10.95 & 	8.17& 	 4.85 	&  \textbf{12.09}
                                                 \\ \hline
PointRCNN       \cite{Shi_2019_CVPR}
&   10.98 & 	\textbf{11.75}& 	 10.20 	&  11.32
                 \\ \hline
PointRCNN-IoU      \cite{Shi_2019_CVPR}
&    10.66 & 	\textbf{12.19}& 	 10.23 & 	 11.30
                         \\ \hline
SECOND     \cite{yan2018second}
 &  \textbf{11.35} & 	8.82& 	 7.46 	&  8.10
                                \\ \hline
SECOND-IoU   \cite{yan2018second}
 &   5.00 & 	\textbf{8.78}& 	 5.23 & 	 7.50
                                 \\ \hline\hline
                                        BEV  (AP@0.50)                             & Center & Line & Pyramid & Trapezoid  \\ \hline
PV-RCNN   \cite{shi2020pv}
   &     15.30  & 	\textbf{19.96} & 	 17.27  & 	 17.23
                    \\ \hline
Voxel RCNN   \cite{deng2020voxel}
&    11.45  & 	9.32 & 	 7.37  & 	 \textbf{13.48}
                                                 \\ \hline
PointRCNN       \cite{Shi_2019_CVPR}
&    11.86  & 	\textbf{14.44} & 	 11.67  & 	 12.21
                 \\ \hline
PointRCNN-IoU    \cite{Shi_2019_CVPR}
&    11.59  & 	\textbf{14.38}	 &  11.85  & 	 13.61
                         \\ \hline
SECOND         \cite{yan2018second}
&    13.38  & 	\textbf{13.71}	 &  10.08  & 	 12.57
                                \\ \hline
SECOND-IoU    \cite{yan2018second}
&   7.46  & 	\textbf{12.27} & 	 8.49  & 	 10.54
                                 \\ \hline
\end{tabular}}
\caption{Influence of LiDAR placement on pedestrian detection}
\vspace*{-3mm}
\end{table}

\subsection{Qualitative Visualization and Analysis}
This section shows some qualitative visualization and analysis of point cloud collected through different LiDAR placements. From Figure \ref{fig:vis1}, we can see that the distribution of point cloud varies a lot in the same scenario, which is directly caused by different LiDAR placements. Specifically, the point distribution of \textit{Center} is more uniform in the vertical direction. However, there are some aggregated points as clear horizontal lines for other placements, which results in the performance improvement for small object detection like Cyclist or extremely large trucks, as shown in Figure \ref{fig:expt2} and Table \ref{tab:expt1}.

 However, since the aggregated lines can better represent the shape and other critical information of the objects,  other plane-placed LiDAR configurations have better performance for large-scale object detection like Car, where there are enough points for the detection task.
In Figure \ref{fig:vis2}, we illustrate the influence of roll and pitch angle in \textit{Pyramid} placement. For the objects far away, the distribution of object point cloud is quite different. The distribution under LiDARs with roll angle is sparser horizontally, making key points harder to aggregate. However, \textit{Pyramid-pitch} placement makes the points in the front object shifted below, resulting in the object not being that clear to detect. Note that there is some slight difference in objects for each scenario, but it does not matter because the distribution of objects is the same for all LiDAR placements. As the number of frames is huge, the detection performance will be fair for all the experiments.

\section{Appendix - Limitation and Discussion}
In this section, we present some limitations of our work and some further discussions. First, when calculating our surrogate metric using POG, we have assumed that all the voxels are independent. Although objects in ROI are independent among all the frames, the voxels may be related to their neighbors and not fully independent. However, since the number of frames is large enough, the related voxels are highly separated, so our assumptions somewhat hold. Besides, we do not consider the case of occlusion. When LiDAR beams meet some object, they will no longer get through it in most cases and reflect. In our Bresenham model, we ignore such occlusion based on observing that the occlusion case is rare in the CARLA town with 40, 80, or 120 vehicles. Also, occlusion often happens far away from the ego vehicle, so it may not influence the surrogate metric too much. Also, we leave the task of finding better surrogate functions to evaluate or even optimize LiDAR configurations as the future work in this community.

Although we have conducted extensive experiments and made comprehensive comparison and analyses, there are still some weaknesses. First, we do not train each model for 80 epochs as default to make the experiment efficient. We have found that  the fine-tuning loss of all the detection algorithms has already converged after 10 epochs,  so we stop it and test the performance for a fair evaluation. Also, the point cloud data is sometimes sparse for objects far away, which is challenging for detection and the detection metrics are not as high as those on KITTI. That is based on the observation that the influence of LiDAR placement is more evident for sparse point cloud to avoid saturation and is also consistent with the practical applications where multiple LiDARs are mainly used towards the challenging cases to detect objects with sparse points. Moreover, the evaluated multi-LiDAR placements are not as complicated as the cases in the company. The real LiDAR placement is confidential for the company, and we believe the simplified case is enough to show the influence of LiDAR configuration to meet the motivation of this work.

\end{document}